\documentclass[letterpaper]{article}
\usepackage[preprint]{aaai2027}
\usepackage[hyphens]{url}
\usepackage{graphicx}
\usepackage{natbib}
\usepackage{caption}
\usepackage{algorithm}
\usepackage{algpseudocode}
\usepackage{amsmath}
\usepackage{amssymb}
\usepackage{array}
\usepackage{colortbl}
\usepackage{xcolor}
\definecolor{KleinBlue}{HTML}{002FA7}
\definecolor{LightGrey}{HTML}{D3D3D3}
\definecolor{aaaihighlight}{HTML}{E1FFFF}
\newcolumntype{L}[1]{>{\raggedright\arraybackslash}p{#1}}
\newcolumntype{C}[1]{>{\centering\arraybackslash}p{#1}}
\newcolumntype{R}[1]{>{\raggedleft\arraybackslash}p{#1}}
\newcommand{\ms}[2]{#1\,{\scriptstyle\pm#2}}

\usepackage{newfloat}
\usepackage{listings}
\usepackage{placeins}
\DeclareCaptionStyle{ruled}{labelfont=normalfont,labelsep=colon,strut=off}
\floatstyle{ruled}
\newfloat{listing}{tb}{lst}{}
\floatname{listing}{Listing}

\usepackage{booktabs}

\title{GVCCTurbo: Rate--Compute Quality Scheduling for Codebook Driven \\ Generative Compression}

\author{
    Ziyue Zeng\textsuperscript{\rm 1},
    Dingjie Peng\textsuperscript{\rm 1},
    Xun Su\textsuperscript{\rm 1},
    Hiroshi Watanabe\textsuperscript{\rm 1}
}

\affiliations{
    \textsuperscript{\rm 1}Waseda University, Tokyo, Japan\\
    zengziyue@fuji.waseda.jp,
    kefipher9013@asagi.waseda.jp,
    suxun\_opt@asagi.waseda.jp,
    hiroshi.watanabe@waseda.jp
}

\begin{document}
\maketitle

\begin{abstract}
Codebook-driven generative compression uses a pretrained image or video generator as a zero-shot visual prior and transmits compact codebook indices to guide reconstruction at ultra-low bitrate. Current codecs tie each finite-rate correction to a fresh prior evaluation, so shortening the sampler also removes correction slots that carry target-dependent information. We propose GVCCTurbo, a BPP-driven scheduler that separates expensive prior refreshes from codebook corrections: after calibrating an atom-count operating point and skip-gap ratio once per protocol, it maps a target codebook-payload bitrate to a trajectory length and refresh period, making BPP a schedule input instead of a fixed consequence of sampler length. The same endpoint-prediction and finite-rate steering interface covers GVCC-style rectified-flow video and DDCM-style diffusion image compression, preserving zero-training deployment and compatibility with future distilled priors. Native 1080p curves position the complete zero-shot codec in the ultra-low-bitrate regime. In a controlled 720p Wan-GVCC study, the scheduler cuts prior evaluations from \(20\) to \(9\) for a \(\sim\!44\%\) measured decoding-time reduction shared across the whole schedule family, at a small shared LPIPS cost on high-motion content; within that family, uniform refresh thinning (pure-skip) is a boundary point, and the BPP-aware interior point trades \(2.9\%\) fewer codebook-payload bits for consistently higher PSNR at comparable LPIPS. These results support BPP-to-compute scheduling as a controllable extension of sampler-length tuning, without requiring the allocated point to dominate every boundary point.
\end{abstract}

\section{Introduction}
\label{sec:intro}

At ultra-low bitrate, the bitstream cannot fully specify visual structure. Conventional hybrid codecs such as HEVC~\cite{sullivan2012overview} and VVC~\cite{bross2021overview}, and learned codecs such as DCVC-FM~\cite{li2024neural} and DCVC-RT~\cite{jia2025dcvcrt}, remain strong at standard bitrates but become deterministic and over-smoothed as fine detail disappears from the transmitted signal. Generative compression addresses this regime by using a pretrained image or video generator as a visual prior, with compact target-dependent bits steering reconstruction toward the input~\cite{mentzer2020hific,theis2022lossy,yang2023lossy,vonderfecht2025lossy,zeng2026gvcc}.

\begin{figure}[!t]
    \centering
    \includegraphics[width=\linewidth]{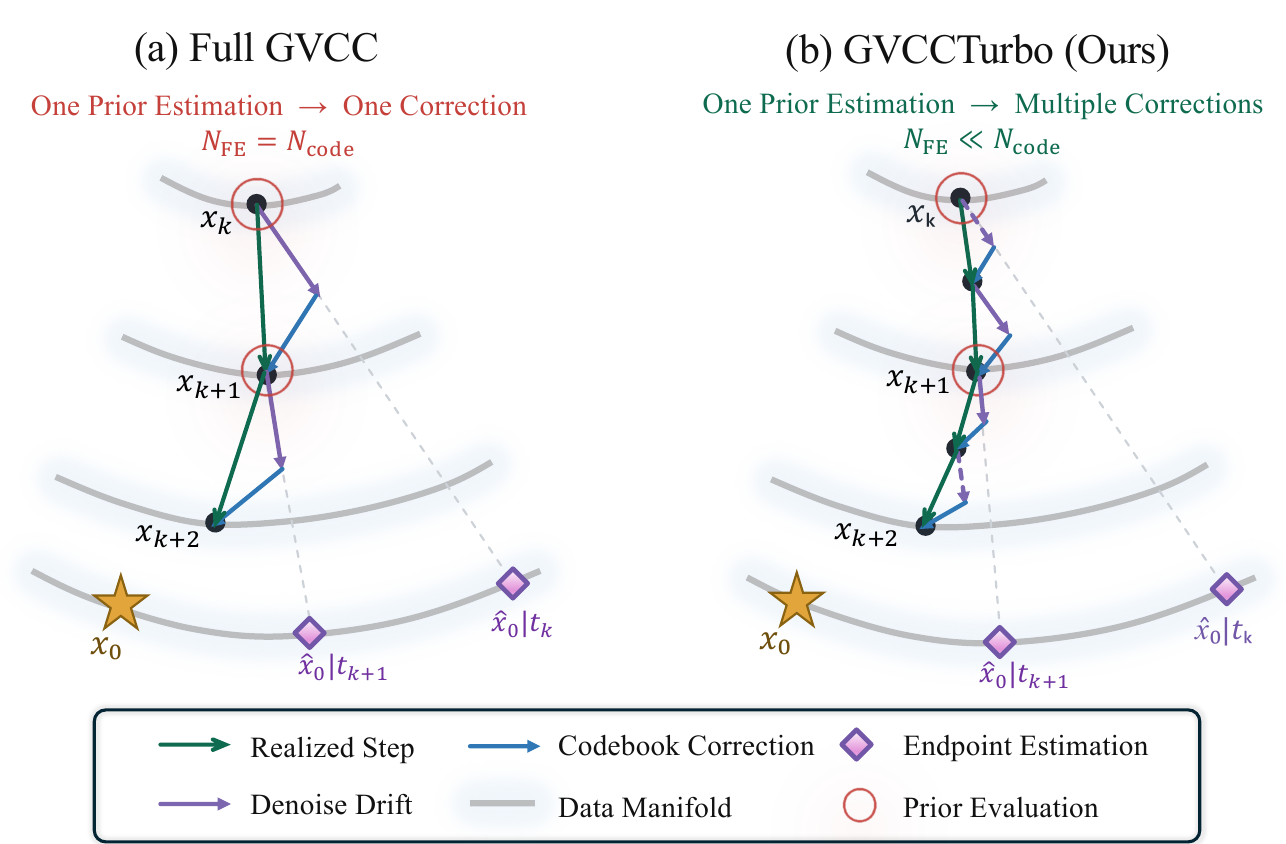}
    \caption{Trajectory geometry. (a)~Full GVCC spends one prior evaluation per correction step; (b)~GVCCTurbo refreshes the prior only at selected steps while keeping the correction grid dense between refreshes.}
    \label{fig:trajectory}
\end{figure}

Generative codecs may use the generator as a perceptually aligned latent space, a reconstruction module conditioned on side information, or---in the narrower model-as-codec class studied here---the codec itself. This distinction matters for acceleration: in the last setting, skipping a generator evaluation changes the bitstream-guided trajectory rather than merely accelerating a downstream module. GVCC~\cite{zeng2026gvcc} uses a pretrained rectified-flow video generator as a zero-shot decoder and transmits compact codebook indices instead of dense residuals; Turbo-DDCM~\cite{vaisman2025turboddcm} develops a related multi-atom mechanism for image diffusion codecs. Both avoid training a task-specific backbone, but tie each correction step to a prior evaluation: shortening the sampler saves computation while removing opportunities to inject target-dependent information.

This coupling is undesirable because bitrate controls finite-rate corrections, whereas compute is dominated by prior evaluations; reconstruction quality depends on both. As illustrated in Fig.~\ref{fig:trajectory}, GVCCTurbo separates them within a \((T,p,M)\) schedule family, with truncation and uniform refresh thinning as boundary points. It keeps the correction grid dense, refreshes the prior only at selected steps, and reuses the cached clean endpoint between refreshes, thereby skipping evaluations without deleting scheduled codebook-payload opportunities. Because encoder-selected innovations make the trajectory non-smooth, skipped steps use an endpoint hold; bit-exact replay further requires a fixed shared schedule rather than online per-sample cache decisions. We calibrate a perceptually stable atom count \(M_{\mathrm{perc}}\) and tolerated skip-gap ratio \(\tau_{\mathrm{gap}}\) once per protocol. At run time, a probe-free calculator fixes \(M=M_{\mathrm{perc}}\), maps a target codebook-payload BPP to \((T^\star,p^\star)\), and returns \((T^\star,p^\star,M_{\mathrm{perc}})\) without retraining the prior.

We make three contributions. First, we separate decoder-reproducible prior evaluations, which add no newly transmitted target bits, from finite-rate corrections carrying target-dependent symbols in the scheduled codebook channel, yielding a common rate--compute interface for GVCC-style rectified-flow video and DDCM-style diffusion image codecs. Second, after one-time protocol calibration of \(M_{\mathrm{perc}}\) and \(\tau_{\mathrm{gap}}\), endpoint-cached refresh and a probe-free calculator map a target codebook-payload BPP to \((T^\star,p^\star,M_{\mathrm{perc}})\) without retraining the prior, codebook, or decoder replay; uniform refresh thinning is a boundary of this schedule family rather than an external baseline. Third, decoupling \(N_{\mathrm{code}}\) from prior-evaluation count removes the structural tie between correction density and sampler length, creating a testable route toward few-evaluation distilled priors rather than establishing distilled-prior gains here; the supplementary material reports only a preliminary observation. Native 1080p curves position the complete codec. Controlled 720p Wan-GVCC studies cut prior evaluations from \(20\) to \(9\), reducing decoding time by \(\sim\!44\%\) across the schedule family with a small shared LPIPS cost on high-motion content; within that family, the BPP-aware point uses \(2.9\%\) fewer codebook-payload bits while improving PSNR at comparable LPIPS. Diagnostics on the FLF2V-14B GVCC backbone and Turbo-DDCM test interface transfer rather than cross-system speed parity.

\section{Related Work}
\label{sec:related}

\subsection{Generative Models in Codecs}
Generative models play several roles in compression~\cite{yang2026survey}. Latent-transform codecs use them to provide perceptually aligned representations while a learned transform and entropy model carry the compressed signal~\cite{qi2025generative,guo2025glvc,zheng2026gvc1d,li2026progvc}; restoration and conditional-generation codecs instead reconstruct detail from a separately coded representation or compact side information~\cite{mao2025gnvcvd,wang2025disco,ding2026cgvc,wang2025tgvc,wan2025m3cvc}. In both cases, accelerating the generator changes a representation or reconstruction module rather than the replay of a transmitted generative trajectory.

Our work is closest to the model-as-codec paradigm, where the bitstream specifies the trajectory of a pretrained generator. DiffC communicates per-step diffusion innovations through reverse channel coding~\cite{theis2022lossy,vonderfecht2025lossy}; GVCC and Turbo-DDCM instead steer rectified-flow video and diffusion image trajectories with reproducible codebooks~\cite{zeng2026gvcc,ohayon2025ddcm,vaisman2025turboddcm}. Related zero-shot approaches use posterior, noise-combination, progressive, or reverse-channel coding~\cite{elata2024psc,su2025ncs,ling2026freegvc}. GVCCTurbo builds on the codebook-correction interface and maps a codebook-payload bitrate cap to trajectory length and refresh period with explicit correction and prior-evaluation costs.

\subsection{Sampling Acceleration}
Distillation, consistency training, and efficient generators reduce sampling cost by changing the model, objective, or representation~\cite{salimans2022progressive,song2023consistency,luo2023latent,xie2024sana}. They are complementary to our zero-training scheduler, which preserves the pretrained prior and bit-exact trajectory replay.

Training-free caches reuse intermediate features or predict model-output changes from timestep and state displacement~\cite{ma2024deepcache,liu2025teacache,ma2025magcache,fan2025taocache,bu2025dicache,cui2026dpcache}. Their smooth, fixed-conditioning trajectories permit extrapolation or online per-sample reuse. Here, encoder-selected codebook innovations perturb every step, favoring a held endpoint over output extrapolation, while bit-exact replay requires a precommitted shared schedule. GVCCTurbo therefore removes prior evaluations symmetrically at the encoder and decoder; it does not claim superiority over decoder-only adaptive caches.

\section{Method}
\label{sec:method}
GVCCTurbo leaves the pretrained generator, shared codebook, and bit-exact replay unchanged, modifying only the shared codec/sampling schedule \((T,p,M)\): trajectory length, prior-refresh period, and per-step atom count. Encoder and decoder derive the same refresh steps, so skipped evaluations require no signaling. Content-adaptive refresh would instead require signaling and, if selected from dense cache metrics, encoder-side probes. This zero-training layer remains compatible with backbone acceleration such as distillation.

\subsection{Unified Interface}
\label{sec:method-interface}

Let \(x_0\in\mathbb{R}^D\) be the target latent of an image or video GOP, and let \(x_k=x_{t_k}\) be the decoder state at reverse timestep \(t_k\). Codebook-driven compression steers a learned prior that predicts the clean target from the current state. At step \(k\), this prediction is
\begin{equation}
D_k(x_k)=\hat{x}_{0|t_k}.
\end{equation}
For a rectified-flow backbone~\cite{liu2022rectified} under the linear-path convention \(x_t=(1-t)x_0+t\epsilon\), the velocity target is \(\epsilon-x_0\), so \(u_\theta(x_t,t)\approx\epsilon-x_0\). The clean endpoint estimate is therefore
\begin{equation}
\hat{x}_{0|t_k}=x_k-t_k u_\theta(x_k,t_k).
\label{eq:x0-hat}
\end{equation}
For diffusion-based codecs~\cite{ho2020ddpm,song2021scoresde}, \(D_k\) denotes the corresponding denoised estimate.

Since the encoder has access to \(x_0\), it computes the endpoint residual
\begin{equation}
r_k=x_0-D_k(x_k).
\label{eq:residual}
\end{equation}
The residual is not transmitted directly. The encoder projects it onto a shared codebook \(C=[c_1,\ldots,c_K]\) that both sides can reproduce, keeping the \(M\) atoms most aligned with \(r_k\):
\begin{equation}
S_k=
\operatorname{TopM}_{j\in\{1,\ldots,K\}}
\left|\langle c_j,r_k\rangle\right|.
\end{equation}
The transmitted innovation is then
\begin{equation}
z_k^\star
=
\operatorname{norm}
\left(
\sum_{j\in S_k}
\operatorname{sign}\!\left(\langle c_j,r_k\rangle\right)c_j
\right),
\label{eq:projection}
\end{equation}
where only the selected indices and signs are encoded, and \(\operatorname{norm}(\cdot)\) matches the sampler-specific stochastic scale.

Projecting the residual into the stochastic slot is not a heuristic choice. Under the linear path \(x_t=(1-t)x_0+t\epsilon\), the same state \(x_k\) admits two decompositions: one with the true pair \((x_0,\epsilon)\), and one with the model estimate and the noise it implicitly attributes, \(x_k=(1-t_k)\hat{x}_{0|t_k}+t_k\hat{\epsilon}_k\). Subtracting the two gives
\begin{equation}
\epsilon-\hat{\epsilon}_k
=
-\frac{1-t_k}{t_k}\,r_k ,
\label{eq:eps-identity}
\end{equation}
so an endpoint error is identically a noise-attribution error: \(r_k\) is, up to a scalar, the innovation missing from the stochastic slot for the target trajectory. The exact missing innovation contains both direction and magnitude. Our fixed-scale codebook channel transmits only a quantized direction, while the admissible correction scale is prescribed by the sampler. Normalization matches this sampler-prescribed scale and avoids a gross scale mismatch with the prior's expected state distribution; because atom selection is conditioned on the target residual, it does not imply an unconditional Gaussian sample. This is analogous to the innovation channel that DiffC~\cite{theis2022lossy,vonderfecht2025lossy} communicates with reverse channel coding; the codebook projection provides a fixed-rate, shared-seed directional quantization. The same identity holds with a path-specific scalar for the DDPM path underlying DDCM, qualifying the endpoint residual as the common interface quantity of both families (full analysis in the supplementary material).

A codebook-controlled update step then takes the form
\begin{equation}
x_{k+1}
=
\mathcal{A}_k\!\left(x_k,D_k(x_k)\right)
+
\mathcal{B}_k z_k^\star,
\label{eq:generic-step}
\end{equation}
where \(\mathcal{A}_k\) is the backbone-specific prior update and \(\mathcal{B}_k z_k^\star\) is the finite-rate correction. Evaluating \(D_k\) is the denoiser or flow-model evaluation counted by \(N_{\mathrm{FE}}\). Applying \(z_k^\star\) spends bitrate and is counted by \(N_{\mathrm{code}}\). The two costs are informationally asymmetric: \(D_k(x_k)\) is a deterministic function of the decoder state and the shared weights, computable without receiving any new bits, whereas every applied \(z_k^\star\) introduces target-dependent symbols in the scheduled codebook channel. Evaluations refine the \emph{anchor} against which residuals are measured; corrections inject the \emph{payload}. Refresh thinning preserves the number of target-dependent correction slots and their nominal rate capacity, while changing the selected innovations through the cached residual. Within the scheduled codebook channel, it changes the anchor used for selection without deleting payload opportunities, so GVCCTurbo schedules the two costs separately.

\begin{figure}[!t]
    \centering
    \includegraphics[width=\linewidth]{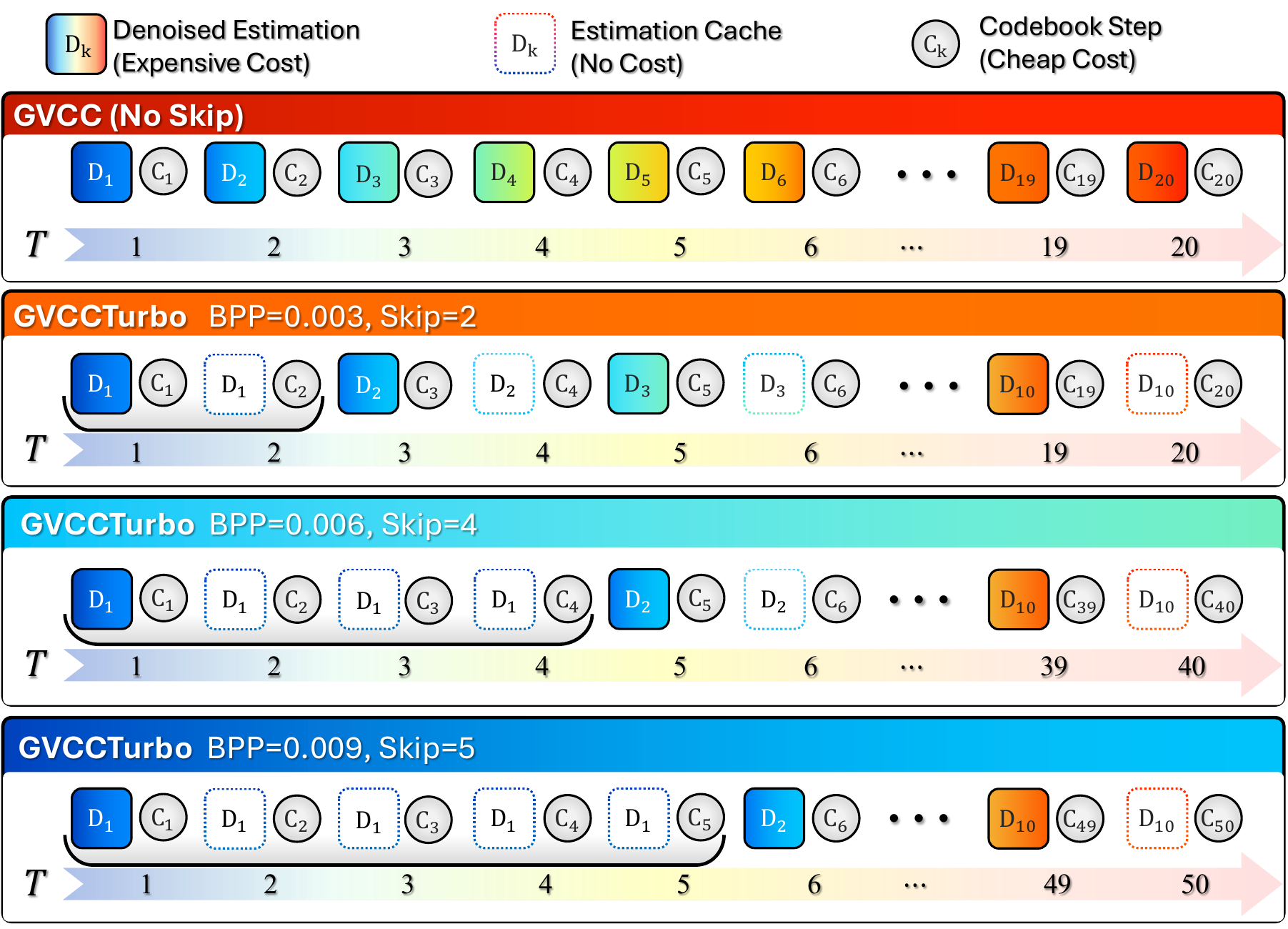}
    \caption{Rate--compute scheduling. GVCC ties each codebook correction to a prior evaluation, whereas GVCCTurbo caches the clean endpoint and reuses it across several corrections, decoupling prior refreshes (\(N_{\mathrm{FE}}\)) from finite-rate corrections (\(N_{\mathrm{code}}\)).}
    \label{fig:scheduling}
\end{figure}

\subsection{Endpoint-Cached Prior Refresh}
\label{sec:method-refresh}

As shown in Fig.~\ref{fig:scheduling}, GVCCTurbo evaluates the prior only at a set of refresh steps and reuses each clean-endpoint prediction across several low-cost codebook corrections. The correction trajectory can therefore remain dense even when the number of prior evaluations is reduced. Let \(\mathcal{R}\) be the set of refresh steps, and let
\(
s(k)=\max\{s\in\mathcal{R}:s\leq k\}
\)
be the most recent refresh step at or before step \(k\). At a refresh step \(s\in\mathcal{R}\), the codec evaluates the prior and stores its clean-endpoint prediction
\begin{equation}
a_s = D_s(x_s)=\hat{x}_{0|t_s}.
\label{eq:cached-endpoint}
\end{equation}
At a non-refresh step \(k\notin\mathcal{R}\), the prior evaluation \(D_k(x_k)\) is skipped, and the cached prediction is reused as
\(\widetilde{D}_k = a_{s(k)}\). The encoder still forms a residual against this cached endpoint, \(\widetilde{r}_k=x_0-\widetilde{D}_k\), selects \(z_k^\star=\Pi_{K,M}(\widetilde{r}_k)\), and transmits its atom indices and signs. Endpoint caching therefore removes prior evaluations without removing any correction step.

For rectified-flow backbones under the linear path convention used in Eq.~\eqref{eq:x0-hat}, the cached endpoint also induces a local velocity at the current state:
\begin{equation}
\widetilde{u}_k
=
\frac{x_k-\widetilde{D}_k}{t_k}.
\label{eq:u-skip}
\end{equation}
The cache therefore stores a predicted endpoint rather than a velocity. Since \(x_k\) changes after each correction step, reusing the endpoint lets the local direction be recomputed from the current state without a new prior evaluation. The held-endpoint velocity component \(x'=(x-a)/t\) is exactly integrable in isolation because \((x-a)/t\) is invariant along it. The full sampler also contains stochastic, score, or correction terms: these use the same numerical scheme but inherit the cached-endpoint approximation. Endpoint staleness, \(e_k=D_k(x_k)-a_{s(k)}\), is therefore the central approximation introduced by reuse, rather than an exact account of the full stochastic update. Under a continuous normalized-direction approximation, small relative staleness causes only second-order angular loss, but the discrete Top-\(M\) selection can change discontinuously; once staleness becomes comparable to the residual, substantial misalignment becomes likely. This local intuition and the measured slow-drift-then-cliff behavior motivate the calibrated skip-gap constraint. The supplementary material gives the scoped analysis.

\subsection{Probe-Free Calibrated Scheduling}
\label{subsec:schedule}

GVCCTurbo separates schedule selection into two codec-level decisions: the strength of each finite-rate correction and the largest protocol-calibrated fraction of the trajectory over which a cached endpoint can be reused. Both quantities are calibrated once for a specified backbone, resolution protocol, and atom-rate model, and are then reused as fixed constants within that setting. At test time the scheduler runs no probe and estimates no per-input statistics; with \(M=M_{\mathrm{perc}}\) fixed, it maps a target codebook-payload BPP directly to \((T^\star,p^\star)\).

\paragraph{Perceptual atom operating point.}
The per-step atom count \(M\) sets both the strength and the bitrate cost of each correction: under a fixed budget, smaller \(M\) buys more correction steps but weakens each projection, and larger \(M\) does the opposite. We select a perceptually stable operating point \(M_{\mathrm{perc}}\) from iso-bitrate and iso-compute sweeps with LPIPS as the primary criterion; the sweep for our Wan-GVCC instantiation is reported in the experiments, and other backbones or atom coders calibrate their own value.

\paragraph{Skip-gap calibration constant.}
Endpoint staleness is the central approximation introduced by reuse. Because it is not available to the decoder without an additional prior evaluation, we use the skipped interval relative to the full controllable trajectory as a protocol-level proxy and calibrate a fixed, input-independent skip-gap ratio \(\tau_{\mathrm{gap}}\) from a fixed-\((T,M)\) cliff sweep. For a trajectory length \(T\), a refresh period \(p\) skips \(p-1\) consecutive prior evaluations, so we take the largest period whose skipped fraction \((p-1)/T\) stays within \(\tau_{\mathrm{gap}}\):
\begin{equation}
p^\star
=
\left\lfloor \tau_{\mathrm{gap}} T \right\rfloor + 1 .
\label{eq:pstar}
\end{equation}
The cliff sweep that fixes \(\tau_{\mathrm{gap}}\) for our instantiation is reported in the ablations; the scheduler then uses \(M_{\mathrm{perc}}\) and \(\tau_{\mathrm{gap}}\) to allocate bitrate and compute, as described next.

\subsection{Rate--Compute Allocation}
\label{sec:method-schedule}

After \(M_{\mathrm{perc}}\) and \(\tau_{\mathrm{gap}}\) are calibrated once for a protocol, GVCCTurbo maps a target codebook-payload BPP \(B\) to \((T^\star,p^\star)\) and returns the shared schedule \((T^\star,p^\star,M_{\mathrm{perc}})\).

Let \(P=FHW\) denote the number of displayed pixels, \(F_{\mathrm{lat}}\) the number of latent temporal slots using codebook corrections, and \(q\) the deterministic tail length with no transmitted codebook bits. For a constant-\(M\) schedule and a generic atom-rate model \(b(M,K)\), the codebook-payload BPP is
\begin{equation}
\operatorname{BPP}_{\mathrm{cb}}(T,M)
=
\frac{
F_{\mathrm{lat}}\,N_{\mathrm{code}}(T)\,b(M,K)
}{P},
\label{eq:bpp-schedule}
\end{equation}
where \(N_{\mathrm{code}}(T)=T-q\). Equation~\eqref{eq:bpp-schedule} counts atom identities and signs, but excludes container headers and non-codebook side information. For a protocol with side information, all terms normalized over the same displayed-pixel support,
\[
\operatorname{BPP}_{\mathrm{tot}}
=
\operatorname{BPP}_{\mathrm{side}}
+
\operatorname{BPP}_{\mathrm{cb}}.
\]
Thus \(B\) in the scheduler denotes the target codebook-payload BPP, not the total codec BPP. Once a possibly content-dependent side rate is known, an external total-rate allocator could set the available codebook budget to \(\max\{0,\operatorname{BPP}_{\mathrm{tot,target}}-\operatorname{BPP}_{\mathrm{side}}\}\); such content-dependent total-rate allocation is not used or claimed in our controlled scheduler experiments, which specify the codebook-payload budget directly. For FLF2V, boundary-frame bits are included separately when reporting total BPP over unique frames. For the signed-index rate model used in Wan-GVCC video, \(b_{\mathrm{vid}}(M,K)=M(\lceil\log_2K\rceil+1)\). For the supplemental Turbo-DDCM image transfer, atom identities are coded as an unordered subset with signs, so \(b_{\mathrm{img}}(M,K)=\lceil\log_2 \binom{K}{M}\rceil+M\). The codebook-rate allocation below is unchanged after substituting the backbone-specific atom-rate model.

With \(M=M_{\mathrm{perc}}\) fixed, the target codebook-payload BPP determines the maximum number of codebook-controlled steps:
\begin{equation}
N_{\mathrm{code}}^\star
=
\left\lfloor
\frac{B\,P}{F_{\mathrm{lat}}\,b(M_{\mathrm{perc}},K)}
\right\rfloor
\label{eq:t-from-bpp}
\end{equation}
This gives as many whole correction steps as the cap allows at the calibrated strength; \(T^\star=N_{\mathrm{code}}^\star+q\), and any residual budget below one correction-step quantum remains unused.

The refresh period is chosen from the fixed skip-gap rule in Eq.~\eqref{eq:pstar}. The resulting number of prior evaluations is
\begin{equation}
N_{\mathrm{FE}}^\star
=
\left\lceil
\frac{N_{\mathrm{code}}^\star}{p^\star}
\right\rceil
+q,
\label{eq:nfe}
\end{equation}
where the two terms count refreshes during codebook-controlled steps and deterministic tail evaluations. The tail length \(q\) counts deterministic DDIM steps with no codebook bits (\(q=3\) video, \(q=1\) image) and enters as a constant offset in \(T\) and \(N_{\mathrm{FE}}\), independent of the bitrate allocation. Compute is thus an output of the BPP-to-schedule rule; an explicit budget acts only as a feasibility constraint \(N_{\mathrm{FE}}^\star\leq B_{\mathrm{FE}}\). Algorithm~1 in the supplementary material summarizes the procedure.

The strict ultra-low profile sets \(B=B_0\), where \(B_0\) is the no-skip codebook-payload anchor. Small-slack profiles use \(B\leq(1+\eta)B_0\) and report \(B/B_0\). Wider BPP sweeps are used only for diagnostics and rate--compute--quality analysis.

\begin{figure*}[!t]
    \centering
    \includegraphics[width=0.9\linewidth]{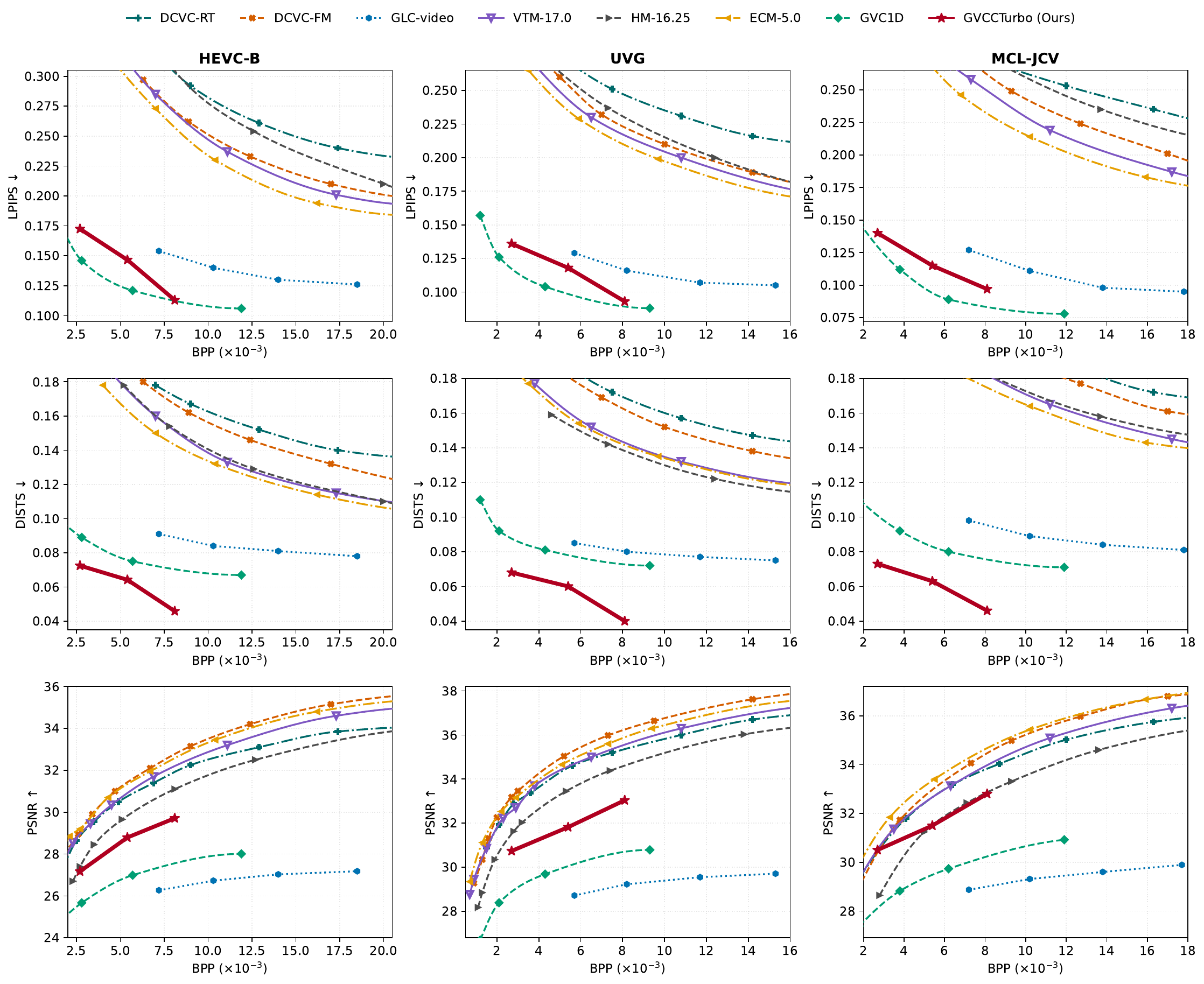}
    \caption{\textbf{Native 1080p rate--distortion comparison} on HEVC-B, UVG,
    and MCL-JCV benchmarks under the same metrics and evaluation protocol.
    BPP is the protocol-total rate.}
    \label{fig:rate-compute}
\end{figure*}

\section{Experiments}
\label{sec:experiments}
We organize the evaluation around three claims: native 1080p codec performance, controlled scheduler behavior, and interface diagnostics across backbones and codec families.

\subsection{Experimental Protocol}
\label{sec:exp-setup}
For the native 1080p comparison we use UVG~\cite{mercat2020uvg}, HEVC Class B, and MCL-JCV~\cite{wang2016mcljcv} with 33-frame GOPs; these system-level points use the FLF2V-14B GVCC backbone with predictive skip and multi-resolution refinement, positioning the complete codec and reporting protocol-total BPP, including boundary-frame bits. For isolation, a controlled 720p study runs Wan-T2V-1.3B~\cite{wan2025} under matched settings, reporting no-skip, pure-skip, and BPP-aware schedules from the same \((T,p,M)\) design family with codebook-payload BPP, \(N_{\mathrm{FE}}\), LPIPS~\cite{zhang2018lpips}, PSNR (DISTS~\cite{ding2020dists} additionally for the native curves), and decoder-only wall-clock reductions. Full protocol details---sequence counts, matched preprocessing and codebook construction, and hardware---are in the supplementary material.

\subsection{Native 1080p Rate--Distortion Curves}
\label{sec:exp-bpp-curves}
Fig.~\ref{fig:rate-compute} provides the native 1080p comparison. Across the three datasets, GVCCTurbo operates at the extreme low-bitrate end of the curves and obtains low LPIPS and DISTS relative to traditional and learned codecs such as HEVC/VVC~\cite{sullivan2012overview,bross2021overview} and the DCVC family~\cite{li2024neural,jia2025dcvcrt}. The most competitive baseline is the trained generative codec GVC1D~\cite{zheng2026gvc1d}, whose tokenizer, transform, and entropy model are optimized specifically for compression. GVCCTurbo requires no codec-specific training and reaches higher PSNR at comparable ultra-low rates, indicating stronger fidelity even where LPIPS favors the trained perceptual-token route. This figure is an R--D comparison under a unified metric protocol, not a decoding-complexity-matched comparison with real-time codecs.

\begin{figure*}[!t]
    \centering
    \includegraphics[width=0.9\linewidth]{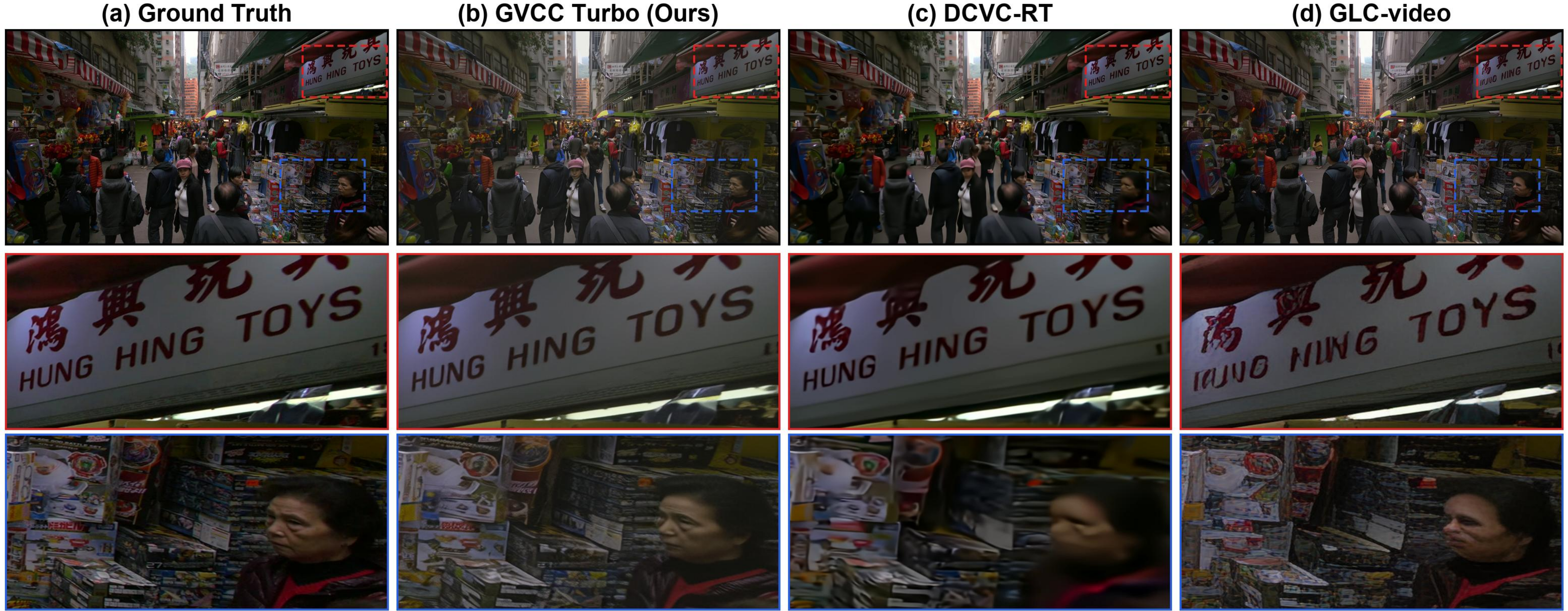}
    \caption{Qualitative 1080p comparison with representative existing
    methods.}
    \label{fig:qualitative}
\end{figure*}

A qualitative 1080p comparison depicted in Fig.~\ref{fig:qualitative} illustrates this behavior: GVCCTurbo preserves the legibility of fine text on the storefront sign and retains coherent facial structure in the cropped region, whereas DCVC-RT over-smooths both regions and GLC-Video introduces visible structural distortion in the text.

\subsection{Controlled Scheduler Study at 720p}
\label{sec:exp-main}

\begin{table}[!t]
\centering
\caption{Ultra-low schedules for Wan-T2V-1.3B at 720p. Here \(B/B_0\) is the codebook-payload BPP ratio. Results are reported as mean \(\pm\) standard deviation, with subscripts indicating \(N_{\mathrm{FE}}\).}
\label{tab:main-ultra-low}
\small
\setlength{\tabcolsep}{2pt}
\begin{tabular}{@{}
  >{\raggedright\arraybackslash}p{1.85cm}
  >{\centering\arraybackslash}p{0.9cm}
  >{\centering\arraybackslash}p{1.7cm}
  >{\centering\arraybackslash}p{1.7cm}
  >{\centering\arraybackslash}p{1.7cm}
@{}}
\toprule
Configuration & \(B/B_0\) & UVG & HEVC-B & MCL-JCV \\
\midrule
\rowcolor{LightGrey}
\multicolumn{5}{@{}l}{\textit{LPIPS} \(\downarrow\) (Primary)} \\
No-skip\(_{20}\)   & 1.000 & \(\ms{0.125}{0.056}\) & \(\ms{0.155}{0.052}\) & \(\ms{0.169}{0.081}\) \\
Pure-skip\(_{9}\)  & 1.000 & \(\ms{0.128}{0.056}\) & \(\ms{0.168}{0.059}\) & \(\ms{0.172}{0.084}\) \\
\rowcolor{aaaihighlight}
GVCCTurbo\(_{9}\)  & 0.971 & \(\ms{0.125}{0.056}\) & \(\ms{0.169}{0.063}\) & \(\ms{0.170}{0.086}\) \\
\addlinespace[1pt]
\(\Delta\)LPIPS & & \(-0.001\) & \(+0.014\) & \(+0.002\) \\
\midrule
\rowcolor{LightGrey}
\multicolumn{5}{@{}l}{\textit{PSNR} \(\uparrow\) (Secondary)} \\
No-skip\(_{20}\)   & 1.000 & \(\ms{28.22}{2.53}\) & \(\ms{25.11}{1.68}\) & \(\ms{26.26}{5.77}\) \\
Pure-skip\(_{9}\)  & 1.000 & \(\ms{28.04}{2.51}\) & \(\ms{24.93}{1.74}\) & \(\ms{26.18}{5.72}\) \\
\rowcolor{aaaihighlight}
GVCCTurbo\(_{9}\)  & 0.971 & \(\ms{28.26}{2.51}\) & \(\ms{25.11}{1.74}\) & \(\ms{26.38}{5.78}\) \\
\midrule
\rowcolor{LightGrey}
\multicolumn{5}{@{}l}{\textit{Decoder-time Reduction} (vs.\ No-skip; shared)} \\
Pure-skip\(_{9}\) & 1.000 & 44\% & 44\% & 44\% \\
\rowcolor{aaaihighlight}
GVCCTurbo\(_{9}\) & 0.971 & 44\% & 44\% & 44\% \\
\bottomrule
\end{tabular}
\end{table}

The controlled study isolates the scheduler under a fixed codec. The no-skip anchor \((20,1,64)\) defines the reference codebook-payload bitrate and quality; pure-skip \((20,3,64)\) is the refresh-thinning boundary that keeps \(T\) and \(M\) fixed while changing \(p\); GVCCTurbo \((25,4,48)\) is the BPP-aware interior allocation, reaching the same \(N_{\mathrm{FE}}=9\) with more correction steps at the protocol-calibrated atom count. That pure-skip alone already recovers the full \(20\!\to\!9\) saving is itself evidence of the separation: the bitstream needs correction density, not prior-evaluation density.

Table~\ref{tab:main-ultra-low} reports the main controlled evidence. Both accelerated schedules cut prior evaluations from \(20\) to \(9\) and share the same \(\sim\!44\%\) decoding-time reduction; the allocated point is marginally slower in absolute time because its five extra correction steps carry a per-step cost independent of \(M\) (supplementary material). Against no-skip, HEVC-B LPIPS rises similarly for both (\(+0.013\) pure-skip, \(+0.014\) allocated), indicating sensitivity to refresh reduction on harder motion rather than an allocation penalty. At matched \(N_{\mathrm{FE}}=9\), the BPP-aware point spends \(2.9\%\) fewer codebook-payload bits on more correction slots at the calibrated atom count; relative to pure-skip its LPIPS changes are \(-0.003/+0.001/-0.002\) and its PSNR is higher on all three datasets. We read these as quality-comparable rate flexibility within the family, not perceptual superiority: the measured reductions substantiate the shared saving rather than ranking the two points. The paired sequence-level mean is \(\Delta\mathrm{LPIPS}=-0.0018\) (\(p=0.046\), \(n=42\)); it characterizes the change but is not used to claim a visible advantage.

\subsection{Schedule-Family Diagnostic}
\label{sec:exp-schedule-family}

\begin{table}[!t]
\centering
\caption{One-time atom-count calibration at fixed \(N_{\mathrm{FE}}=9\) and near-\(B_0\) codebook-payload bitrate.}
\label{tab:m-select}
\small
\setlength{\tabcolsep}{5pt}
\begin{tabular}{@{}cccccl@{}}
\toprule
\(M\) & \((T,p)\) & \(N_{\mathrm{code}}\) & LPIPS \(\downarrow\) & PSNR \(\uparrow\) & Role \\
\midrule
12 & (93,16) & 90 & 0.198 & \textbf{28.34} & Weak Atoms \\
18 & (63,10) & 60 & 0.195 & 28.28 & Weak Atoms \\
27 & (43,7)  & 40 & 0.173 & 28.27 & Over-long \\
40 & (30,5)  & 27 & 0.160 & 27.99 & Near Optimum \\
\rowcolor{aaaihighlight}

48 & (25,4)  & 22 & \textbf{0.156} & 27.96 & Selected \\
64 & (20,3)  & 17 & 0.159 & 27.62 & Pure-skip \\
\bottomrule
\end{tabular}
\end{table}

Table~\ref{tab:m-select} diagnoses the schedule family at fixed compute, \(N_{\mathrm{FE}}=9\), and near-fixed bitrate (all rows within \(3\%\) of \(B_0\), most within \(1\%\)). Small atom counts spread weaker corrections over long trajectories, while \(M=64\) recovers the pure-skip boundary. The literal truncation corner (\(p=1\)) would force a short \(T=9\) trajectory with only six coarse correction slots, so it is a structural boundary, not a measured row. Endpoint caching expands the feasible space: pure-skip keeps seventeen correction slots, and the allocated schedule keeps twenty-two by using the protocol-calibrated \(M_{\mathrm{perc}}=48\). Because the one-GOP LPIPS separation is small, the table is used to select \(M_{\mathrm{perc}}\) and illustrate schedule freedom, not to rank family members; the skip-gap sweep and the calculator's behavior across bitrate targets appear in the ablations and supplementary material.

\subsection{Cross-Backbone Transfer}
\label{sec:exp-flf2v}
The FLF2V-14B GVCC backbone tests whether the same scheduling interface transfers to a stronger, boundary-conditioned model, whose bitrate splits into a scheduled codebook component and a boundary-frame component computed over unique frames.

\begin{figure}[!t]
    \centering
    \includegraphics[width=0.8\linewidth]{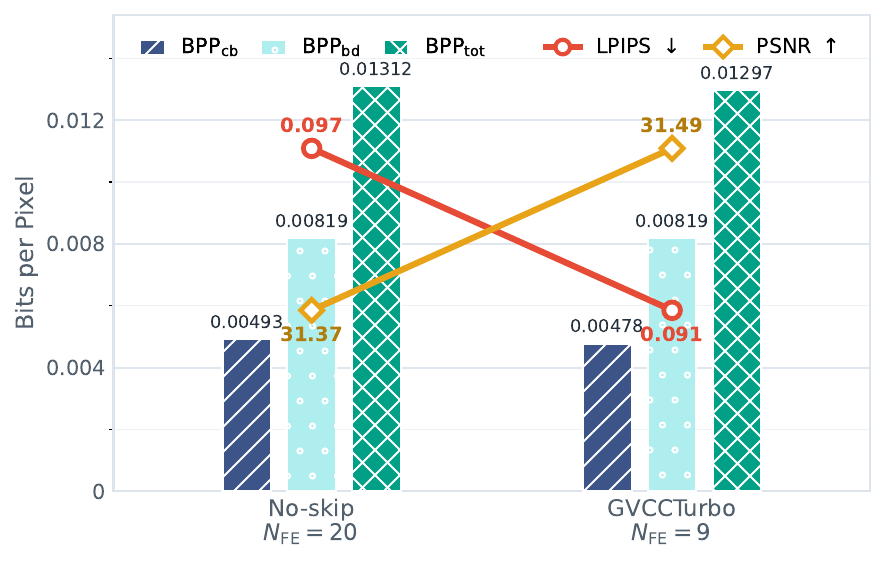}
    \caption{Cross-backbone transfer to FLF2V-14B on UVG.}
\label{fig:flf2v-transfer}
\end{figure}

As shown in Fig.~\ref{fig:flf2v-transfer}, on this 14B backbone the same interface reduces \(N_{\mathrm{FE}}\) from \(20\) to \(9\) while matching the no-skip anchor in LPIPS and PSNR at 720p. Because boundary-frame bits dominate, codebook savings barely change total BPP; this experiment therefore tests cross-backbone NFE scheduling, not BPP gain. The supplementary material includes the matched pure-skip endpoint. Turbo-DDCM with SD-2.1 provides a second-codec interface check: pure-skip gives LPIPS \(0.099\) at \(1.000B_0\) and \(N_{\mathrm{FE}}=16\), whereas the BPP-aware point gives \(0.096\) at \(0.991B_0\) and \(N_{\mathrm{FE}}=15\). We do not treat this small difference as a powered quality or speed claim; in the small image latent, \(N_{\mathrm{FE}}\) no longer tracks wall-clock time (supplementary material).

% \begin{table}[!t]
% \centering
% \caption{Cross-backbone transfer to FLF2V-14B on UVG at 720p.}
% \label{tab:flf2v-transfer}
% \small
% \setlength{\tabcolsep}{1pt}
% \begin{tabular}{@{}L{0.24\linewidth}C{0.07\linewidth}C{0.14\linewidth}C{0.14\linewidth}C{0.14\linewidth}C{0.09\linewidth}C{0.09\linewidth}@{}}
% \toprule
% Method & \(N_{\mathrm{FE}}\) & \(\mathrm{BPP}_{\mathrm{cb}}\) & \(\mathrm{BPP}_{\mathrm{bd}}\) & \(\mathrm{BPP}_{\mathrm{tot}}\) & LPIPS \(\downarrow\) & PSNR \(\uparrow\) \\
% \midrule
% No-skip     & 20 & 0.00493 & 0.00819 & 0.01312 & 0.097 & 31.37 \\
% GVCCTurbo   & 9  & 0.00478 & 0.00819 & 0.01297 & 0.091 & 31.49 \\
% \bottomrule
% \end{tabular}
% \end{table}

\section{Ablation Study}
\label{sec:ablations}
The ablations characterize three parts of the scheduler: (i) how the \((T,p,M)\) family expands sampler-length control, (ii) how the calibrated constants are chosen, and (iii) why the cached quantity is the clean endpoint.

\paragraph{Calibrated Allocation over the Schedule Family.}
Table~\ref{tab:m-select} sweeps the iso-\(N_{\mathrm{FE}}\) design family over \(M\). The LPIPS curve is not monotone, and the stable operating region appears near \(M=48\), placing refresh thinning as one boundary of a larger BPP-planning space. This one-time sweep fixes \(M_{\mathrm{perc}}\); at run time, GVCCTurbo holds \(M=M_{\mathrm{perc}}\), allocates \(N_{\mathrm{code}}\) from the target codebook-payload BPP, and makes the induced \(N_{\mathrm{FE}}\) explicit.

\paragraph{Fixed Skip-Gap Rather than Per-Input Probing.}
Table~\ref{tab:skip-gap-ablation} calibrates the second constant, \(\tau_{\mathrm{gap}}\), using a fixed-\((T,M)\) stress probe. A conservative period \(p=4\) uses more prior evaluations, while \(p=7\) reaches \(N_{\mathrm{FE}}=9\) with a skipped fraction of \(14.0\%\). Increasing the period beyond the measured band degrades quality: LPIPS reaches \(0.304\) at \(p=9\), and the reconstruction collapses at \(p=15\). Textured and smooth diagnostics, together with five held-out clips and a multi-GOP variance check in the supplement, show a compatible stability region within the calibrated Wan-GVCC 720p protocol. This supports the protocol-fixed choice \(\tau_{\mathrm{gap}}\approx0.15\), used by all controlled experiments, without implying invariance across resolutions or GOP settings. Because LPIPS already drifts from \(0.244\) (\(p{=}4\)) to \(0.273\) (\(p{=}7\)), \(\tau_{\mathrm{gap}}\) marks a stability bound, not a no-degradation region.

\paragraph{Endpoint Caching compared with Velocity Caching.}
GVCCTurbo caches the clean endpoint prediction \(\hat{x}_0\) rather than the instantaneous velocity. A frozen velocity cannot adapt to the updated state after subsequent prior drift and codebook innovations, whereas an endpoint cache allows the local velocity to be recomputed from the current state through Eq.~\eqref{eq:u-skip}. A controlled comparison with this simple frozen-velocity baseline confirms the advantage of endpoint caching. At the same \(M_{\mathrm{perc}}=48\) operating point (\(N_{\mathrm{FE}}=9\), one-GOP diagnostic) with bit-exact-identical codebook and seed, frozen-velocity reuse degrades LPIPS from \(0.156\) to \(0.257\) (\(+65\%\), worse on all seven sequences) and PSNR from \(27.96\) to \(25.42\)~dB at the same \(N_{\mathrm{FE}}\) and bitrate; per-sequence values are reported in the supplementary material. Higher-order extrapolators could in principle reduce cache staleness further, but would introduce additional state and scheduling choices beyond the present fixed-schedule interface.
\begin{table}[!t]
\centering
\caption{Skip-gap ablation at fixed \(T=43\) and \(M=27\) on a two-sequence stress subset.}

\label{tab:skip-gap-ablation}
\small
\begin{tabular}{@{}
  >{\centering\arraybackslash}p{1.2cm}
  >{\centering\arraybackslash}p{1.4cm}
  >{\centering\arraybackslash}p{1.cm}
  >{\centering\arraybackslash}p{1.5cm}
  >{\centering\arraybackslash}p{1.5cm}
@{}}
\toprule
\(p\) & gap/\(T\) & \(N_{\mathrm{FE}}\) & LPIPS \(\downarrow\) & PSNR \(\uparrow\) \\
\midrule
4  & 7.0\%  & 13 & 0.244 & 26.16 \\
\rowcolor{aaaihighlight}
7  & 14.0\% & 9  & 0.273 & 26.09 \\
9  & 18.6\% & 8  & 0.304 & 25.80 \\
15 & 32.6\% & 6  & 0.565 & 23.27 \\
\bottomrule
\end{tabular}
\end{table}

\section{Conclusion}
\label{sec:conclusion}
GVCCTurbo separates prior evaluations from finite-rate corrections and schedules them through a shared endpoint-cache interface without changing the pretrained prior, codebook, or deterministic replay. In the controlled 720p study, prior evaluations fall from \(20\) to \(9\), and decoding time falls by \(\sim\!44\%\); within this accelerated family, the BPP-aware point uses \(2.9\%\) fewer codebook-payload bits at comparable quality. The current constants are calibrated per protocol and are not shown resolution- or GOP-invariant; temporal metrics are not measured, and \(N_{\mathrm{FE}}\) tracks wall-clock only when prior evaluations dominate. Reported timings are decoder-only and do not imply parity with trained real-time codecs. Because the interface is training-free and backbone-agnostic, future few-step distilled priors can be tested by replacing the prior and recalibrating the schedule.

\bibliography{aaai2027}

\clearpage
\appendix
\setcounter{secnumdepth}{2}
\section{Experiment Setup}
\label{app:setup}

\paragraph{Implementation details.}
Unless otherwise specified, we use codebook size \(K=16384\), base atom count \(M_0=64\), base trajectory length \(T_0=20\), deterministic tail length \(q=3\), SDE diffusion coefficient \(g_{\mathrm{scale}}=3.0\), classifier-free guidance scale \(1.0\), timestep shift \(5.0\), and fixed seed \(42\). The native 1080p comparison uses UVG (7 sequences), HEVC Class B (5 sequences), and the full MCL-JCV set (30 sequences); the controlled 720p study shares matched preprocessing, GOP partitioning, codebook construction, and backbone settings. Prior-evaluation counts fall symmetrically at the encoder and decoder; the wall-clock reductions reported in the main paper are decoder-only. Each 720p sequence uses GOP length 33 and latent shape \(16\times9\times90\times160\), with the number of GOPs per sequence adjusted to scene changes; per-dataset scores are sequence-equal-weighted means. Codebook bitrate is counted as \(M(\lceil\log_2K\rceil+1)\) bits per latent slot. Experiments run on Ubuntu~24.04 with PyTorch~2.10 (video) and PyTorch~2.11 (image transfer) under CUDA~12.8; timings are measured on a single NVIDIA RTX PRO 6000 (96\,GB) under identical precision and batch settings. Both the native 1080p rate--distortion curves and the 720p controlled study use the full MCL-JCV dataset (videoSRC01--30).

\paragraph{Metrics.}
BPP is computed as \(B_{\mathrm{total}}/(FHW)\), where \(B_{\mathrm{total}}\) is the sum of the bitrate components reported by the relevant protocol. The controlled Wan-GVCC and Turbo-DDCM scheduler studies count the scheduled codebook payload; FLF2V additionally includes boundary-frame bits over unique frames. Container headers are excluded. PSNR is computed from RGB MSE after mapping pixels to \([0,1]\), and LPIPS uses the AlexNet backbone with inputs mapped to \([-1,1]\).

\section{Mechanistic Intuition for Thinning Prior Refreshes}
\label{app:mechanism}

The main paper reports that the schedule removes roughly half of the prior
evaluations at small perceptual cost, and calibrates the stability region
empirically through \(\tau_{\mathrm{gap}}\).  This appendix gives the
interface-level reasons for that robustness.  The arguments use the
endpoint/noise identity under the appropriate path convention and the
codebook projection shared by the GVCC and DDCM instances.  The
rectified-flow derivation is given first, followed by its DDPM counterpart.
Notation follows the main
paper: \(x_0\) is the target latent, \(D_k(x_k)=\hat{x}_{0|t_k}\) the
clean-endpoint prediction at step \(k\), \(r_k=x_0-D_k(x_k)\) the residual,
\(a_{s(k)}\) the cached endpoint from the most recent refresh step, and
\(z_k^\star\) the transmitted signed-atom innovation.

\subsection{Prior Evaluations Add No New Transmitted Bits}
\label{app:mech-blind}

A prior evaluation \(D_k(x_k)\) is a deterministic function of the decoder
state and the shared weights, so the decoder computes it without receiving
new bits.  Because the state already contains past target-dependent
corrections, however, the evaluation need not be statistically independent
of the target.  Within the scheduled codebook channel, newly communicated
target-dependent symbols enter through the finite-rate corrections:
\(N_{\mathrm{code}}\) selections of \(M\) signed atoms.  Removing evaluations
while retaining the correction count and atom count therefore preserves the
number of correction slots and their nominal capacity, but it does not leave
the bitstream content unchanged: the cached residual can select different
indices and signs.  Evaluations and corrections remain asymmetric resources
because the former refine the anchor without new transmitted symbols,
whereas the latter spend the scheduled codebook payload.

\subsection{The Residual Is Mis-Attributed Noise}
\label{app:mech-epsilon}

Why does projecting the endpoint residual into the sampler's
\emph{stochastic} slot steer the trajectory at all?  Under the linear path
\(x_t=(1-t)x_0+t\epsilon\), the same state \(x_k\) admits two
decompositions: one with the true pair \((x_0,\epsilon_{\mathrm{true}})\)
and one with the model's estimate \(\hat{x}_{0|t_k}\) and the noise
\(\hat{\epsilon}_k\) it implicitly attributes,
\(x_k=(1-t_k)\hat{x}_{0|t_k}+t_k\hat{\epsilon}_k\).  Subtracting the two
gives
\begin{equation}
\epsilon_{\mathrm{true}}-\hat{\epsilon}_k
=
-\frac{1-t_k}{t_k}\,r_k .
\label{eq:eps-identity-supp}
\end{equation}
An endpoint error is thus \emph{identically} a noise-attribution error, up
to a scalar factor.  The exact missing innovation contains both direction
and magnitude.  The codec imposes a fixed-scale channel design: it transmits
a finite-rate quantization of the direction, while the admissible correction
scale is prescribed by the sampler.  Normalization matches that prescribed
scale and avoids a gross scale mismatch with the prior's expected state
distribution.  It does not imply an unconditional Gaussian sample, because
the Top-\(M\) atoms and their signs are selected conditionally on the target
residual.
The identity is not specific to rectified flow: for the DDPM path
\(x_t=\sqrt{\bar{\alpha}_t}\,x_0+\sqrt{1-\bar{\alpha}_t}\,\epsilon\)
underlying DDCM, the same two-decomposition subtraction gives
\(\epsilon-\hat{\epsilon}_k
=-\sqrt{\bar{\alpha}_t/(1-\bar{\alpha}_t)}\;r_k\).
The endpoint residual is therefore the mis-attributed innovation in both
families, up to a path-specific scalar, which is what qualifies it as the
common quantity of the unified interface in the main paper.  This channel
view originates with DiffC~\cite{theis2022lossy,vonderfecht2025lossy},
which transmits an approximate posterior sample of the per-step innovation
by reverse channel coding at substantial computational cost; the codebook
projection can be read as a fixed-rate, shared-seed quantization of the
corresponding missing-innovation direction.

\subsection{Exact Drift under a Held Endpoint}
\label{app:mech-drift}

Endpoint caching replaces \(D_k(x_k)\) by a held prediction \(a\) and the
induced velocity \(\widetilde{u}_k=(x_k-a)/t_k\).  In isolation, the
resulting drift component \(x'(t)=(x-a)/t\) satisfies
\begin{equation}
\frac{d}{dt}\!\left[\frac{x-a}{t}\right]
=
\frac{x'(t)\,t-(x-a)}{t^{2}}
=0,
\label{eq:drift-invariant}
\end{equation}
so \((x-a)/t\) is invariant and the state moves along the straight segment
toward the cached endpoint.  Sub-steps that recompute
\(\widetilde{u}_k\) from the current state reproduce the exact solution of
this held-endpoint component.  The full sampler also contains stochastic,
score, or correction terms.  Although these use the same numerical scheme,
their values inherit the cached-endpoint approximation, so the exactness
above does not extend to the full stochastic update.  Endpoint staleness---
the gap between \(a_{s(k)}\) and what \(D_k(x_k)\) would have returned---is
the central approximation introduced by reuse.  Each stochastic correction
still has its own prescribed step scale and fresh codebook draw, which
motivates thinning \(N_{\mathrm{FE}}\) while keeping
\(N_{\mathrm{code}}\) dense.

\subsection{Staleness Explains the Empirical Cliff}
\label{app:mech-staleness}

Between refreshes the encoder measures the frozen residual
\(\widetilde{r}_k=x_0-a_{s(k)}\), whereas the true residual is
\(r_k=x_0-D_k(x_k)\).  Their gap is the staleness
\(e_k=D_k(x_k)-a_{s(k)}\), giving \(\widetilde{r}_k=r_k+e_k\).  While
\(\|e_k\|\ll\|\widetilde{r}_k\|\), a continuous normalized-direction
approximation gives an angular loss of order
\(O(\|e_k\|^2/\|\widetilde{r}_k\|^2)\).  This is local intuition rather than
a guarantee: discrete Top-\(M\) selections can change discontinuously even
under a small perturbation.  As the skipped interval grows and the two norms
become comparable, substantial misalignment becomes likely.  This predicts
a gentle local regime followed by a sharper degradation and is consistent
with the measured slow-drift-then-cliff shape used to calibrate
\(\tau_{\mathrm{gap}}\); its numerical value remains protocol-calibrated.
The same accounting also explains why holding the endpoint can outperform
freezing the velocity (main-paper ablation): a frozen velocity cannot adapt
to the updated state, whereas an endpoint cache allows the local velocity to
be recomputed from the current state.

\subsection{Why Dense Corrections Need Not Be Redundant}
\label{app:mech-window}

Within a skip window, all corrections target the same frozen
\(\widetilde{r}\), and each step draws a fresh shared-seed codebook.  The
selected innovations are conditioned on the same residual, so they are
neither independent nor unconditional Gaussian samples.  Nevertheless,
fresh draws make exact repetition unlikely and vary the orthogonal
quantization components, providing an intuition for why dense corrections
need not collapse into duplicates.  A simple high-dimensional
order-statistic approximation places the top-\(M\) threshold among \(K\)
atoms at order \(\sqrt{2\log(K/M)}\).  Comparing the coherently accumulated
component along \(\widetilde{r}\) with the incoherent orthogonal component
then gives the rough alignment scale
\begin{equation}
\cos\!\big(z^\star,\widetilde{r}\big)
\;\approx\;
\sqrt{\frac{2M\log(K/M)}{D}} .
\label{eq:alignment}
\end{equation}
Equation~\eqref{eq:alignment} is an order approximation that neglects
selection dependencies, normalization details, and closed-loop dynamics.
Under the signed-index video rate model, the directional gain grows
sublinearly with \(M\) while bit cost grows linearly; other atom coders use
their own \(b(M,K)\).  This offers intuition for an interior operating point
but neither proves complementarity nor ranks schedules.  The actual
\(M_{\mathrm{perc}}\) is therefore calibrated empirically.

\section{Toward Distilled Priors: An Implication and a Preliminary Observation}
\label{app:distillation}

A potential implication of decoupling the correction count
\(N_{\mathrm{code}}\) from the prior-evaluation count \(N_{\mathrm{FE}}\) is
that a dense correction grid can coexist with few prior evaluations. This
appendix explains why tying the two is restrictive for a few-step prior and
reports a preliminary observation. It motivates future study rather than
establishing a distilled-prior gain or a universal necessity.

\paragraph{Why tying is restrictive for a distilled prior.}
A distilled prior is trained to be accurate within a small native
prior-evaluation budget \(B_{\mathrm{FE}}\) (few-step sampling). In a codec
that ties one correction to every prior evaluation---the setting of GVCC and
Turbo-DDCM before this work---a distilled prior faces a tradeoff. To keep
a dense correction grid it must be evaluated many times, which pushes it far
outside its few-step training regime and lets the endpoint prediction drift, so
reconstruction degrades. To instead stay near its native regime it must use few
evaluations, which under tying also means few corrections, limiting the
target-dependent symbols available through the scheduled codebook channel
(Appendix~\ref{app:mechanism}).
Endpoint-cached refresh relaxes this coupling: it evaluates the prior only
\(B_{\mathrm{FE}}\) times---keeping it inside its native regime---while still
applying a correction at every trajectory step, so the correction grid stays
dense. The interface therefore suggests that decoupling may recover some of
the quality that tying forfeits when the prior is distilled.

\paragraph{Preliminary observation.}
A two-sequence single-GOP diagnostic on a step-distilled backbone is consistent
with this prediction. Applying the distilled prior in the tied regime---one
correction per evaluation along a long trajectory---produces markedly degraded
reconstructions, as the prior is driven well outside its few-step operating
point. Endpoint-cached refresh, which invokes the distilled prior only at sparse
refresh steps while preserving the dense correction grid, recovers a large
fraction of this loss at a fraction of the prior evaluations. We report this as
a qualitative phenomenon rather than a measured claim: the diagnostic is small,
uses constants that would be recalibrated for a distilled backbone, and---
crucially---the recovery mitigates the tying-induced collapse rather than
matching full-prior quality, which remains a separate property of the distilled
backbone and carries its own distillation penalty.

\paragraph{Outlook.}
The observation suggests that the decoupling introduced here may make it
easier to use distilled priors inside a codebook-driven codec, and it turns
the distillation--inheritance argument into a concrete test: a
controlled tied-versus-decoupled comparison at matched \(N_{\mathrm{FE}}\),
across distilled backbones and step budgets, with the two calibrated constants
re-measured per backbone. Such a study would determine whether decoupling
materially helps once each backbone is recalibrated; failure to recover
quality would limit the practical value of the interface in that setting. We
leave this full study to future work, as it depends on the availability of
distilled video priors that accept codebook steering.

\section{Scheduler Diagnostics and Supplemental Transfers}
\label{app:scheduler-diagnostics}

This appendix complements the main-paper ablations, which already report the
\(M\)-sweep and skip-gap sweep that fix \(M_{\mathrm{perc}}\approx48\) and
\(\tau_{\mathrm{gap}}\approx0.15\). Here we give the scheduler pseudocode, the
full calculator schedules (Table~\ref{tab:calculator-schedules}) and rate--distortion curve across bitrate targets (Table~\ref{tab:rd-validation}), and
the cross-backbone and cross-codec transfer diagnostics. These diagnostics
focus on the BPP-to-schedule interface itself: the codec backbone and decoder
replay remain fixed, while the one-time protocol calibration fixes
\(M_{\mathrm{perc}}\) and the calculator changes \((T,p)\) under the
appropriate atom-rate model and calibrated constants.

\subsection{Scheduler Pseudocode}

\begin{algorithm}[H]
\caption{Rate--Compute Allocation}
\label{alg:schedule}
\footnotesize
\begin{algorithmic}[1]
\Require Target codebook-payload BPP \(B\); pixel count \(P\); latent slots \(F_{\mathrm{lat}}\); codebook size \(K\); tail length \(q\); atom-rate model \(b(M,K)\); calibrated atom count \(M_{\mathrm{perc}}\); skip-gap constant \(\tau_{\mathrm{gap}}\).
\Ensure Schedule \((T^\star,p^\star,M_{\mathrm{perc}})\), correction count \(N_{\mathrm{code}}^\star\), and prior-evaluation count \(N_{\mathrm{FE}}^\star\).

\Statex \textbf{Rate allocation:}
\State \(b^\star \gets b(M_{\mathrm{perc}},K)\)
\State \(N_{\mathrm{code}}^\star \gets \lfloor B\,P/(F_{\mathrm{lat}} b^\star)\rfloor\)
\State \(T^\star \gets N_{\mathrm{code}}^\star+q\)

\Statex \textbf{Prior refresh:}
\State \(p^\star \gets \lfloor \tau_{\mathrm{gap}} T^\star\rfloor+1\)
\State \(N_{\mathrm{FE}}^\star \gets \lceil N_{\mathrm{code}}^\star/p^\star\rceil+q\)

\Statex \textbf{Schedule construction:}
\For{\(k=0,\ldots,N_{\mathrm{code}}^\star-1\)}
    \If{\(k \bmod p^\star = 0\)}
        \State refresh \(D_k(x_k)\) and cache endpoint \(a_k\)
        \State \(\widetilde{r}_k \gets x_0-a_k\)
    \Else
        \State reuse cached endpoint \(a_{s(k)}\)
        \State \(\widetilde{r}_k \gets x_0-a_{s(k)}\)
    \EndIf
    \State select and apply \(z_k^\star=\Pi_{K,M_{\mathrm{perc}}}(\widetilde{r}_k)\)
\EndFor
\State run \(q\) deterministic tail steps

\State \Return \((T^\star,p^\star,M_{\mathrm{perc}},N_{\mathrm{code}}^\star,N_{\mathrm{FE}}^\star)\)
\end{algorithmic}
\end{algorithm}

\subsection{Held-out Constant Calibration}

As a robustness check, we rerun both sweeps on held-out material disjoint from every benchmark: three stock-market clips, one animation sequence, and one synthetic screen-content clip, spanning natural, animated, and screen domains, one GOP each. As shown in Table~\ref{tab:heldout-calibration}, the sweeps recover the same operating region: the mean iso-\(N_{\mathrm{FE}}\) \(M\)-sweep is minimized at \(M\!=\!48\) on a broad \(40\)--\(64\) plateau (LPIPS within \(0.002\) across the plateau), and the skip-gap curve stays in a gentle-degradation regime (\(\approx\!+0.015\) LPIPS per step up to \(18.6\%\)) before collapsing at \(32.6\%\). Per-clip argmins vary within the plateau---the screen-content clip, whose absolute LPIPS is only \(0.048\), prefers small \(M\), while others favor \(48\) or \(64\)---but the cross-domain mean remains at \(M\!=\!48\), supporting reuse of the same constants in the reported Wan-GVCC 720p protocol.

\begin{table}[H]
\centering
\caption{\textbf{Out-of-distribution held-out calibration} on five clips outside all evaluation benchmarks (three stock-market, one animation, one synthetic screen-content; one GOP each). (a) iso-\(N_{\mathrm{FE}}=9\), iso-\(B_0\) \(M\)-sweep, mean LPIPS; (b) fixed \((T,M)=(43,27)\) \(p\)-sweep, mean LPIPS. The mean optimum reproduces \(M_{\mathrm{perc}}\!\approx\!48\) and the \(\tau_{\mathrm{gap}}\!\approx\!0.15\) cliff. Absolute LPIPS is not comparable to the in-distribution schedule-selection and skip-gap tables of the main paper.}
\label{tab:heldout-calibration}
\small
\setlength{\tabcolsep}{5pt}
\begin{tabular}{@{}lcccccc@{}}
\multicolumn{7}{@{}l}{(a) \(M\)-sweep (iso-\(N_{\mathrm{FE}}=9\))} \\
\toprule
\(M\) & 12 & 18 & 27 & 40 & 48 & 64 \\
\midrule
LPIPS \(\downarrow\) & 0.173 & 0.175 & 0.160 & 0.157 & \textbf{0.155} & 0.157 \\
\bottomrule
\end{tabular}

\vspace{0.6em}
\begin{tabular}{@{}lccccc@{}}
\multicolumn{6}{@{}l}{(b) \(p\)-sweep (fixed \(T{=}43,M{=}27\))} \\
\toprule
\(p\) & 1 & 4 & 7 & 9 & 15 \\
gap/\(T\) & 0\% & 7.0\% & 14.0\% & 18.6\% & 32.6\% \\
\midrule
LPIPS \(\downarrow\) & 0.131 & 0.143 & 0.160 & 0.175 & 0.279 \\
\bottomrule
\end{tabular}
\end{table}

\paragraph{Multi-GOP variance of the calibration sweeps.}
The held-out check above varies \emph{content}; we additionally vary the \emph{time segment} by re-running both sweeps over three consecutive 33-frame GOPs per sequence under the same diagnostic protocol. The \(M\)-sweep uses the seven diagnostic sequences (20 sequence--GOP cells; one clip is only two GOPs long), and the skip-gap sweep uses four sequences---the original textured/smooth stress pair plus HoneyBee and videoSRC07 (12 cells). As shown in Table~\ref{tab:multigop-variance}, the mean \(M\)-sweep is again minimized at \(M{=}48\) with the \(40\)--\(64\) plateau within \(0.003\) LPIPS, and per-cell argmins concentrate on \(40\)--\(48\) (15/20 cells; the remainder fall to \(27\), never to the weak-atom or pure-skip extremes). The skip-gap curve degrades gently up to an \(18.6\%\) skipped fraction (largest per-cell \(p{=}7\!\to\!9\) increase \(+0.031\) LPIPS) and collapses at \(32.6\%\) in \emph{all} 12 cells (smallest collapse jump \(+0.058\)), so the cliff location does not drift across sequences or GOPs. The per-cell standard deviations below reflect content difficulty, not instability of the constants.

\begin{table}[H]
\centering
\caption{\textbf{Multi-GOP variance of both calibration sweeps} (three GOPs per sequence, same protocol as the main-paper diagnostics; mean \(\pm\) std across sequence--GOP cells). (a) iso-\(N_{\mathrm{FE}}=9\), iso-\(B_0\) \(M\)-sweep over the seven diagnostic sequences (20 cells). (b) fixed \((T,M)=(43,27)\) \(p\)-sweep over four sequences (12 cells). The optimum stays at \(M_{\mathrm{perc}}{=}48\) and the skip-gap cliff stays between \(18.6\%\) and \(32.6\%\), matching the single-GOP calibration and the held-out check above.}
\label{tab:multigop-variance}
\small
\setlength{\tabcolsep}{3pt}
\begin{tabular}{@{}lcccccc@{}}
\multicolumn{7}{@{}l}{(a) \(M\)-sweep (iso-\(N_{\mathrm{FE}}=9\)), 20 sequence--GOP cells} \\
\toprule
\(M\) & 12 & 18 & 27 & 40 & 48 & 64 \\
\midrule
mean LPIPS \(\downarrow\) & 0.186 & 0.183 & 0.162 & 0.152 & \textbf{0.149} & 0.152 \\
std & 0.112 & 0.108 & 0.091 & 0.078 & 0.076 & 0.075 \\
\bottomrule
\end{tabular}

\vspace{0.6em}
\begin{tabular}{@{}lccccc@{}}
\multicolumn{6}{@{}l}{(b) \(p\)-sweep (fixed \(T{=}43,M{=}27\)), 12 sequence--GOP cells} \\
\toprule
\(p\) & 1 & 4 & 7 & 9 & 15 \\
gap/\(T\) & 0\% & 7.0\% & 14.0\% & 18.6\% & 32.6\% \\
\midrule
mean LPIPS \(\downarrow\) & 0.138 & 0.148 & 0.163 & 0.177 & 0.297 \\
std & 0.077 & 0.085 & 0.096 & 0.104 & 0.177 \\
\bottomrule
\end{tabular}
\end{table}

\subsection{Supplemental Codebook-BPP Operating Points}

\paragraph{Robustness beyond the Strict Anchor.}
Within the same calibrated protocol, the calculator maps additional codebook-payload BPP targets
to schedules while holding \(N_{\mathrm{FE}}=9\). Across UVG, HEVC-B, and
MCL-JCV, these settings cut decoding time by \(43\%\)--\(45\%\), consistent
with the \(\sim\!44\%\) reduction at the strict anchor, while improving the
rate--distortion point. At \(1.191B_0\), the three-dataset scores are LPIPS
\(0.120/0.160/0.165\) and PSNR \(28.66/25.46/26.73\), respectively; the added
correction steps carry only a small \(M\)-independent per-step cost. The
scheduler traces a local, protocol-specific frontier, not a universal one.

\begin{table}[H]
\centering
\caption{\textbf{Calculator schedules implied by the codebook-payload bitrate cap.}  \(M\) stays
at the perceptual near-optimum (\(M_{\mathrm{perc}}=48\)); extra bits buy
correction steps, and the calibrated skip-gap rule holds \(N_{\mathrm{FE}}\) near 9.  Here
\(\mathrm{gap}/T=(p^\star-1)/T\) is the \emph{realized} skipped fraction, kept
just under \(\tau_{\mathrm{gap}}=0.15\).}
\label{tab:calculator-schedules}
\small
\setlength{\tabcolsep}{1.5pt}
\begin{tabular}{@{}L{0.16\linewidth}C{0.17\linewidth}C{0.14\linewidth}C{0.13\linewidth}C{0.11\linewidth}C{0.07\linewidth}@{}}
\toprule
Target & \((T,p,M)\) & \(N_{\mathrm{code}}\) & \(\mathrm{BPP}_{\mathrm{cb}}/B_0\) & gap/\(T\) & \(N_{\mathrm{FE}}\) \\
\midrule
\(B_0\) & (25,4,48) & 22 & 0.971 & 12.0\% & 9 \\
\(1.2B_0\) & (30,5,48) & 27 & 1.191 & 13.3\% & 9 \\
\(1.3B_0\) & (32,5,48) & 29 & 1.279 & 12.5\% & 9 \\
\bottomrule
\end{tabular}
\end{table}

\begin{figure}[H]
\centering
\includegraphics[width=\linewidth]{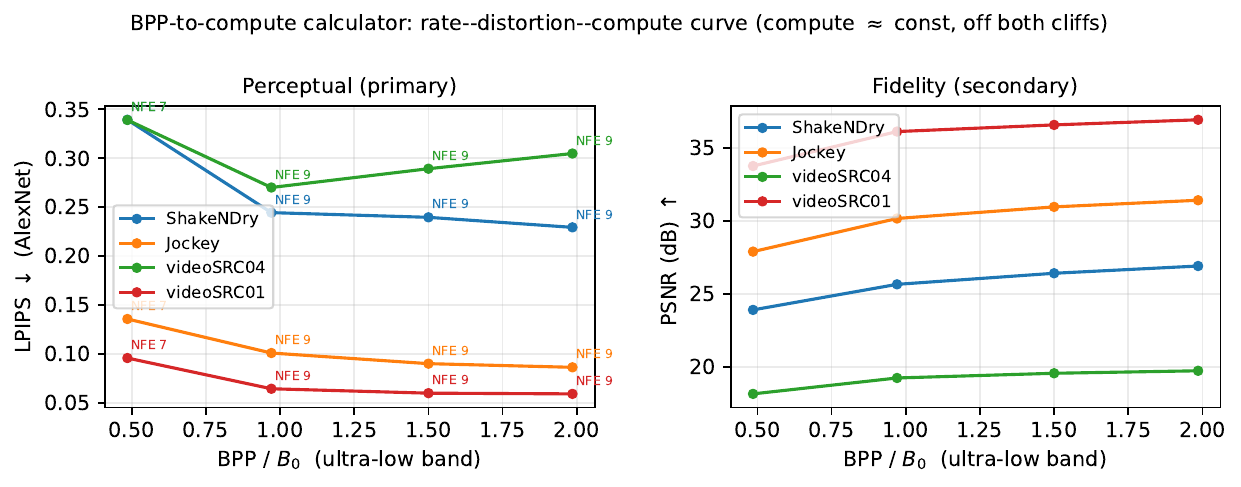}
\caption{\textbf{Probe-free codebook-payload bitrate curve at \(M_{\mathrm{perc}}{=}48\) on a
one-GOP diagnostic subset.}  Two sequences spanning textured-hard (ShakeNDry)
to smooth (videoSRC01) share the calculator's schedules at each bitrate target.
This figure supplements the fixed-\(B_0\), three-dataset controlled study in the main paper.}
\label{fig:rd-curve}
\end{figure}

\begin{table}[H]
\centering
\caption{\textbf{Probe-free R-D curve at \(M_{\mathrm{perc}}{=}48\) (one GOP;
cf.\ Fig.~\ref{fig:rd-curve}).}
The calculator maps each codebook-payload bitrate target to a schedule; a textured (ShakeNDry)
and a smooth (videoSRC01) sequence are shown.  Increasing codebook-payload BPP improves LPIPS
while \(N_{\mathrm{FE}}\) stays nearly flat (\(7\)--\(9\)).  The
\(\mathrm{BPP}_{\mathrm{cb}}/B_0\)
column lists the \emph{target} ratio; realized ratios follow
Table~\ref{tab:calculator-schedules} (e.g., \(0.971\) at the \(B_0\) target).}
\label{tab:rd-validation}
\small
\setlength{\tabcolsep}{1.5pt}
\begin{tabular}{@{}L{0.24\linewidth}C{0.11\linewidth}C{0.18\linewidth}C{0.08\linewidth}C{0.11\linewidth}C{0.11\linewidth}@{}}
\toprule
Content & \(\mathrm{BPP}_{\mathrm{cb}}/B_0\) & \((T,p,M)\) & \(N_{\mathrm{FE}}\) & LPIPS & PSNR \\
\midrule
ShakeNDry & 0.50 & (14,3,48) & 7 & 0.339 & 23.90 \\
ShakeNDry & 1.00 & (25,4,48) & 9 & 0.244 & 25.65 \\
ShakeNDry & 1.50 & (37,6,48) & 9 & 0.240 & 26.41 \\
ShakeNDry & 2.00 & (48,8,48) & 9 & 0.229 & 26.90 \\
\midrule
videoSRC01 & 0.50 & (14,3,48) & 7 & 0.096 & 33.75 \\
videoSRC01 & 1.00 & (25,4,48) & 9 & 0.064 & 36.11 \\
videoSRC01 & 1.50 & (37,6,48) & 9 & 0.060 & 36.57 \\
videoSRC01 & 2.00 & (48,8,48) & 9 & 0.059 & 36.92 \\
\bottomrule
\end{tabular}
\end{table}

\subsection{Endpoint and Transfer Diagnostics}

Table~\ref{tab:velocity-ablation} gives the per-sequence endpoint-vs-velocity comparison summarized in the main-paper ablation. Both arms share the same \(M_{\mathrm{perc}}=48\) schedule (\(N_{\mathrm{FE}}=9\)), codebook, and seed bit-exactly, and differ only in the cached quantity. Caching the velocity degrades LPIPS on all seven sequences (mean \(0.156\rightarrow0.257\), \(+65\%\)) and PSNR by \(2.5\)~dB, confirming that the endpoint, not the direction, is the quantity to reuse.

\begin{table}[H]
\centering
\caption{\textbf{Endpoint vs.\ velocity caching (per-sequence, one-GOP).}
Both arms use the same \(M_{\mathrm{perc}}=48\) schedule (\(N_{\mathrm{FE}}=9\)) with bit-exact-identical codebook and seed; GVCCTurbo stores the clean endpoint \(\hat{x}_0\) and recomputes the local direction, whereas the velocity arm is a simple frozen-velocity baseline that reuses \(u_\theta(x_s,t_s)\). Lower LPIPS is better.}
\label{tab:velocity-ablation}
\small
\setlength{\tabcolsep}{6pt}
\begin{tabular}{@{}lcc@{}}
\toprule
Sequence & Endpoint \(\hat{x}_0\) & Velocity \(u_\theta\) \\
\midrule
videoSRC01 & 0.065 & 0.174 \\
HoneyBee   & 0.068 & 0.118 \\
Jockey     & 0.099 & 0.169 \\
YachtRide  & 0.115 & 0.178 \\
videoSRC07 & 0.223 & 0.352 \\
ShakeNDry  & 0.244 & 0.356 \\
videoSRC04 & 0.274 & 0.454 \\
\midrule
Mean       & \textbf{0.156} & 0.257 \\
\bottomrule
\end{tabular}
\end{table}

Table~\ref{tab:flf2v-transfer-diagnostic} reports the FLF2V-14B schedule-family diagnostic on UVG 720p, where pure-skip is the \(M{=}64\) refresh-thinning endpoint of the schedule family.

\begin{table}[H]
\centering
\caption{\textbf{FLF2V-14B schedule-family diagnostic (UVG 720p).}  Pure-skip is
the \(M=64\) endpoint of the schedule family.}
\label{tab:flf2v-transfer-diagnostic}
\small
\setlength{\tabcolsep}{1.5pt}
\begin{tabular}{@{}L{0.22\linewidth}C{0.07\linewidth}C{0.14\linewidth}C{0.14\linewidth}C{0.14\linewidth}C{0.09\linewidth}C{0.09\linewidth}@{}}
\toprule
Method & \(N_{\mathrm{FE}}\) & \(\mathrm{BPP}_{\mathrm{cb}}\) & \(\mathrm{BPP}_{\mathrm{bd}}\) & \(\mathrm{BPP}_{\mathrm{tot}}\) & LPIPS & PSNR \\
\midrule
No-skip     & 20 & 0.00493 & 0.00819 & 0.01312 & 0.097 & 31.37 \\
Pure-skip   & 9  & 0.00493 & 0.00819 & 0.01312 & 0.094 & 31.35 \\
GVCCTurbo   & 9  & 0.00478 & 0.00819 & 0.01297 & 0.091 & 31.49 \\
\bottomrule
\end{tabular}
\end{table}

\begin{table}[H]
\centering
\caption{\textbf{Rate-allocation and endpoint-cache interface check on Turbo-DDCM image compression.}
Results are averaged over 5 Kodak images at \(512^2\) with SD-2.1.
All settings use the same no-skip codebook-payload reference bitrate \(B_0\!\approx\!0.108\) bpp.
The image branch uses tail length \(q=1\) (the video branch uses \(q=3\)); the \(N_{\mathrm{FE}}\) column is computed with this \(q\).
\(\Delta\)LPIPS is relative to no-skip. The pure-skip and BPP-aware columns are
two points in the same schedule family. This second-codec instance deliberately
enters the small-latent regime where per-step correction cost is non-negligible, so it
maps the boundary of the \(N_{\mathrm{FE}}\) compute proxy (see text) rather than
restating the video runtime result.}
\label{tab:image-ddcm}
\small
\setlength{\tabcolsep}{4pt}
\begin{tabular}{@{}lccc@{}}
\toprule
 & No-skip ref. & Pure-skip & GVCCTurbo \\
\midrule
\((T,p,M)\) & \((30,1,100)\) & \((30,2,100)\) & \((97,7,25)\) \\
\(N_{\mathrm{FE}}\) & 30 & 16 & 15 \\
\(\mathrm{BPP}_{\mathrm{cb}}/B_0\) & 1.000 & 1.000 & 0.991 \\
LPIPS \(\downarrow\) & 0.090 & 0.099 & 0.096 \\
\(\Delta\)LPIPS \(\downarrow\) & -- & +0.009 & +0.006 \\
PSNR \(\uparrow\) & 25.57 & 25.48 & 25.50 \\
Dec. red. & 0\% & 38\% & 25\% \\
\bottomrule
\end{tabular}
\end{table}

For this image diagnostic, substituting the Turbo-DDCM atom-rate model carries
over the rate-allocation part of the calculator: with \(K=16384\), an
\(M\)-atom correction costs
\(\lceil\log_2 \binom{K}{M}\rceil+M\) bits, replacing the signed-index video
cost.  We fix the image operating point at \(M=25\) before evaluation; the
codebook budget then gives \(N_{\mathrm{code}}=96\) and \(T=97\).  To place
the small-latent diagnostic near the matched compute target
\(N_{\mathrm{FE}}=15\), we set \(p=7\).  Thus this experiment transfers the
rate-allocation and endpoint-cache interfaces, not the Wan-specific
\(\tau_{\mathrm{gap}}\) calibration rule.  It shows that those interfaces can
be instantiated with a second codec and a different atom-rate model without
changing the codec training procedure.

This instance tests the boundary of the \(N_{\mathrm{FE}}\) compute proxy.  The controlled scheduler study established that a per-step
codebook correction carries a real cost independent of the atom count \(M\), so
\(N_{\mathrm{FE}}\) tracks wall-clock only while that per-step cost stays small
next to a prior evaluation.  The image setting deliberately inverts this ratio:
in the small SD-2.1 latent a prior evaluation is cheap, and the BPP-aware
\(T=97\) schedule deliberately buys \(67\) additional correction steps relative
to the pure-skip boundary.  It therefore spends one fewer prior evaluation but
has a smaller wall-clock reduction (\(25\%\) versus \(38\%\) over no-skip).  The interface
and calculator carry over, while compute accounting must move from an
\(N_{\mathrm{FE}}\) count to a per-step-aware model once corrections stop being
negligible. This matches the main-paper limitation that \(N_{\mathrm{FE}}\) tracks wall-clock only when prior evaluations dominate.

\section{1080p Comparison}
\label{app:external}

This appendix gives the extended 1080p comparison referenced in the native
1080p discussion.  It complements the main three-dataset BPP curves with a
tiered tabular comparison.  The table groups codecs by bitrate regime and uses
the same UVG benchmark, preprocessing, and PSNR/LPIPS/BPP metric protocol as
the main 1080p comparison.  These system-level BPP values are protocol-total
rates; the FLF2V values include both codebook and boundary-frame components.
All baselines are open-source implementations
evaluated under this unified protocol; some recent trained generative codecs
report stronger numbers but release neither code nor bitstreams, so they
cannot be evaluated under matched preprocessing and metrics and are omitted.
The comparison therefore represents the reproducible open-source state of the
art rather than every published number.  The controlled 720p T2V-1.3B protocol remains the
mechanism validation for the probe-free scheduler constants.

\begin{table*}[t]
\centering
\caption{\textbf{Additional comparison on UVG 1080p.} Methods are grouped into
bitrate tiers and reported with the same PSNR, LPIPS, and BPP columns used in
the native 1080p comparison. LPIPS remains the primary perceptual metric in the
controlled scheduler protocol. We
report only the FLF2V native-resolution GVCCTurbo point, because the updated
first/last-frame pipeline is the relevant high-resolution variant; T2V and I2V
GVCCTurbo quality rows are intentionally omitted.  The
\textsc{GVCCTurbo-FLF2V} 1080p point uses the multi-resolution{+}skip variant
described in the MR appendix.}
\label{tab:external-main}
\small
\setlength{\tabcolsep}{3pt}
\begin{tabular}{@{}l ccc ccc ccc@{}}
\toprule
 & \multicolumn{3}{c}{Tier 1: $\sim$0.003\,bpp} & \multicolumn{3}{c}{Tier 2: $\sim$0.006\,bpp} & \multicolumn{3}{c@{}}{Tier 3: $\sim$0.05\,bpp} \\
\cmidrule(lr){2-4} \cmidrule(lr){5-7} \cmidrule(l){8-10}
Method & PSNR$\uparrow$ & LPIPS$\downarrow$ & BPP & PSNR$\uparrow$ & LPIPS$\downarrow$ & BPP & PSNR$\uparrow$ & LPIPS$\downarrow$ & BPP \\
\midrule
\multicolumn{10}{@{}l}{\textit{Traditional}} \\
HEVC~\cite{sullivan2012overview} & 29.8 & 0.385 & 0.003 & 30.2 & 0.360 & 0.006 & 34.7 & 0.115 & 0.050 \\
VVC~\cite{bross2021overview}     & 31.5 & 0.325 & 0.003 & 31.8 & 0.315 & 0.006 & 36.0 & 0.112 & 0.050 \\
\midrule
\multicolumn{10}{@{}l}{\textit{Learned}} \\
DCVC-FM~\cite{li2024neural}    & 34.0 & 0.286 & 0.003 & 34.7 & 0.265 & 0.006 & 37.0 & 0.110 & 0.050 \\
DCVC-RT~\cite{jia2025dcvcrt}   & 34.1 & 0.285 & 0.003 & 34.6 & 0.261 & 0.006 & 39.7 & 0.115 & 0.050 \\
\midrule
\multicolumn{10}{@{}l}{\textit{Generative (trained)}} \\
GLC-Video~\cite{qi2025generative} & 27.5 & 0.160 & 0.003 & 29.5 & 0.145 & 0.006 & 32.0 & 0.109 & 0.050 \\
GNVC-VD~\cite{mao2025gnvcvd}     & 26.8 & 0.165 & 0.003 & 30.0 & 0.140 & 0.006 & --- & --- & --- \\
\midrule
\multicolumn{10}{@{}l}{\textit{Generative (zero-shot)}} \\
Free-GVC~\cite{ling2026freegvc}      & --- & 0.185 & 0.003 & --- & 0.155 & 0.006 & --- & 0.105 & 0.050 \\
GVCC-FLF2V~\cite{zeng2026gvcc}       & & & & 31.7 & 0.117 & 0.006 & & & \\
\rowcolor{aaaihighlight}
\textbf{GVCCTurbo-FLF2V} (14B)          & & & & 32.0 & 0.103 & 0.006 & & & \\
\bottomrule
\end{tabular}
\end{table*}

In the \(\sim\!0.006\) tier of Table~\ref{tab:external-main}, \textsc{GVCCTurbo-FLF2V} has lower LPIPS
than GVCC-FLF2V (\(0.103\) vs.\ \(0.117\)). The learned DCVC codecs
reach higher PSNR but have roughly \(2.5\times\) higher LPIPS in this tier. Relative runtime and
multi-resolution acceleration details appear in the MR appendix.  The
representative qualitative comparison is reported in the main paper.

\section{Multi-Resolution GVCC}
\label{app:mr}

The MR branch shares the main experimental setup (Appendix~\ref{app:setup}): \(M_{\mathrm{perc}}\approx48\) for the Wan-GVCC 720p setting, with the no-skip anchor \((T,p,M)=(20,1,64)\) at codebook-payload reference bitrate \(B_0\approx0.00483\) bpp.

The main paper accelerates codebook-driven generative compression along the
trajectory-time axis, evaluating the prior sparsely while keeping finite-rate
corrections dense.  Multi-resolution (MR) GVCC operates on an orthogonal axis,
the resolution axis.  It runs the early high-noise trajectory steps on a smaller
latent grid, restores the native grid near the reconstruction stage, and then
continues with the same codebook-driven decode.  We invoke this branch only for
the native 1080p comparison runs.  It is not part of the 720p controlled
scheduler ablation.

\subsection{Relation to Predictive Skip}
\label{app:mr-relation}

Predictive skip changes how often the expensive prior is evaluated while leaving
the codebook-correction trajectory dense.  MR instead changes the spatial token
count of the early prior evaluations.  The two axes are compatible but separate:
\[
\begin{aligned}
\text{time axis:}\quad &N_{\mathrm{FE}}\neq N_{\mathrm{code}},\\
\text{resolution axis:}\quad &(H_{\mathrm{low}},W_{\mathrm{low}})
  \rightarrow (H,W).
\end{aligned}
\]
The 1080p \textsc{GVCCTurbo-FLF2V} point combines
MR with predictive skip.  Skip supplies the acceleration relative to GVCC at
matched backbone quality, and MR acts as a high-resolution refinement knob that
improves the native 1080p operating point.  MR is absent from the 720p
T2V-1.3B main tables, and we make no claim that it transfers to images.

\subsection{Progressive Latent Resolution}
\label{app:mr-transition}

The video branch follows a progressive latent path.  Early steps run on a
lower-resolution latent grid, and at a chosen transition step the decoder opens
the full grid and fills the newly exposed high-frequency band.  The low-stage
target is not produced by independently encoding a downsampled video.  It is the
low-frequency projection of the native latent,
\begin{equation}
x_0^\text{low}=\mathcal{T}_L(\mathrm{VAE}_\text{full}(V)),
\label{eq:lp-raw-app}
\end{equation}
so that the low and high stages share one coordinate system.  This ``lp\_raw''
convention removes the mismatch that would otherwise arise between the two
stages.

Random spectral expansion leaves the newly opened band untargeted.  The
compression-specific DLC variant adds a transition codebook in which the encoder
approximates the target high-frequency band with signed atoms and transmits
their indices.  The high-band fill is
\begin{equation}
x_{\tau}^{H}=t_{\tau}\epsilon_H+(1-t_{\tau})z_H^\star ,
\label{eq:dlc1-codebook}
\end{equation}
where \(H\) denotes the newly opened high-frequency band, \(\epsilon_H\) is a
decoder-reproducible stochastic initializer, and \(z_H^\star\) is the
target-aware transition correction.  Following the transition, the decoder runs
the remaining full-resolution trajectory steps and can apply the same
endpoint-cache skip schedule used in the main method.

\subsection{Video MR Evidence}
\label{app:mr-results}

\begin{table}[t]
\centering
\caption{Multi-resolution GVCC on the full UVG-7 set (Wan-T2V-1.3B, 720p,
\(480p\!\rightarrow\!720p\); 7 sequences \(\times\) 3 GOPs).  BPP ratio and
decoder speedup are relative to full-resolution GVCC on the same platform, not
absolute-time claims.}
\label{tab:mr-fill-ablation}
\small
\setlength{\tabcolsep}{1.5pt}
\begin{tabular}{@{}L{0.25\linewidth}C{0.16\linewidth}C{0.12\linewidth}C{0.11\linewidth}C{0.13\linewidth}C{0.11\linewidth}@{}}
\toprule
Method & Transition & PSNR & LPIPS & BPP ratio & Dec. speed \\
\midrule
Full-res GVCC & none & 28.72 & 0.114 & 1.000$\times$ & 1.00$\times$ \\
MR-random & random & 27.67 & 0.135 & 0.938$\times$ & 1.38$\times$ \\
\rowcolor{aaaihighlight}
MR-codebook & codebook & 28.07 & 0.122 & 0.996$\times$ & 1.38$\times$ \\
\bottomrule
\end{tabular}
\end{table}

The transition codebook (Table~\ref{tab:mr-fill-ablation}) improves on random high-frequency expansion, closing
roughly \(40\%\) of the MR-random-to-full-resolution PSNR gap (\(+0.40\) dB,
\(-0.012\) LPIPS at the same \(1.38\times\) decoder speedup), while preserving
the bit-exact decoder interface of the main method.  A residual \(-0.65\) dB gap
to full-resolution GVCC remains, which is why MR stays an appendix speed-quality
knob rather than a part of the main GVCCTurbo contribution.

\subsection{Why Image-Domain MR Was Rejected}
\label{app:mr-image-negative}

We also tested a direct two-stage MR port to image-domain Turbo-DDCM
(SD2.1-base); the results are reported in Table~\ref{tab:mr-image-negative}.  The setup mirrors the video idea: the first \(T_\text{low}\)
bit-spending steps run on a \(4\times32\times32\) latent, spectrally expand to
the native \(4\times64\times64\) latent, and then finish the remaining
bit-spending and clean DDIM steps at full resolution.  The baseline is the
single-resolution Turbo-DDCM point \(T=30,M=100\), at PSNR \(27.05\), BPP
\(0.0781\), and \(30\) prior evaluations (\(N_{\mathrm{FE}}\)).

\begin{table}[t]
\centering
\caption{Image-domain MR port on Turbo-DDCM (Kodak pilot diagnostic, SD2.1-base).  The
best case is a near tie, and more low-resolution steps monotonically lose
PSNR.  This result is used only to delimit the scope of MR.}
\label{tab:mr-image-negative}
\small
\setlength{\tabcolsep}{3pt}
\begin{tabular}{@{}C{0.16\linewidth}C{0.21\linewidth}C{0.17\linewidth}C{0.20\linewidth}@{}}
\toprule
\(T_\text{low}\) & PSNR, align & PSNR, none & Best \(\Delta\) \\
\midrule
3  & 27.01 & 27.03 & \(-0.02\) \\
5  & 26.93 & 27.03 & \(-0.02\) \\
7  & 26.49 & 26.66 & \(-0.39\) \\
10 & 23.86 & 25.77 & \(-1.29\) \\
13 & 20.93 & 24.62 & \(-2.43\) \\
16 & 17.59 & 23.90 & \(-3.16\) \\
21 & 15.19 & 23.15 & \(-3.91\) \\
\bottomrule
\end{tabular}
\end{table}

Three causes explain this negative result.  First, image latents are already
small: the native image latent has dimension \(4\cdot64\cdot64=16{,}384\),
whereas a 720p video GOP latent is orders of magnitude larger, so the saved image
UNet work does not offset the transition bookkeeping.  Second, SD2.1-base is
trained on the native \(64\times64\) latent grid, and running the prior at
\(32\times32\) introduces an out-of-distribution low-stage trajectory that the
late full-resolution stage cannot fully repair.  Third, the image early-noise
region is milder than the video one: the measured image spectrum is roughly
\(\omega^{-1.92}\), against \(\omega^{-2.42}\) for the video branch.  MR
therefore helps only when the high-resolution video trajectory carries enough
token count and early-frequency asymmetry to exploit.  For images, the canonical
extension remains the BPP-to-compute scheduler of the main paper rather than MR.

\section{Pixel-Decoder Extension: PiD-DDCM}
\label{app:pid}

PiD-DDCM is a decoder-side perceptual extension.  Where the main paper schedules
prior evaluations and codebook corrections ahead of decoding, PiD addresses a
separate question: once a low-bitrate DDCM/Turbo-DDCM latent has been
transmitted, can a pixel-diffusion decoder yield a more perceptually plausible
4K reconstruction than a standard cascade super-resolution decoder?  We treat
this appendix as a demonstration and diagnostic study; it does not contribute to
the main rate--distortion or speed claims.

\subsection{Decoder-Side 4K Pipeline}
\label{app:pid-purpose}

The compressed representation stays a codebook-driven latent stream, and only
the final image decoder changes.  The tested 4K pipeline is
\[
\begin{aligned}
&4096^2\ \text{GT}
\rightarrow 1024^2\ \text{encoder input}\\
&\rightarrow \text{DDCM-FLUX bits}\\
&\rightarrow
\begin{cases}
\text{VAE decode at }1024^2\ \text{then bicubic/Lanczos }4\times,\\
\text{PiD pixel-diffusion }4\times\ \text{decode}.
\end{cases}
\end{aligned}
\]
PiD is a four-step FLUX-distilled pixel-diffusion SR decoder operating at the
native \(2\mathrm{K}\rightarrow4\mathrm{K}\) point.  The DDCM configuration uses
\(T=28\), codebook size \(K=16384\), \(M=256\), and 21 bit-carrying steps.
We report 4K BPP as the bit count normalized by the \(4096^2\) output pixels,
\begin{equation}
\mathrm{BPP}_{4K}
  = \mathrm{BPP}_{1024}\left(\frac{1024}{4096}\right)^2 .
\label{eq:pid-bpp4k}
\end{equation}
Early-exit variants hand an intermediate latent \(x_{k_e}\) to PiD after only a
prefix of the DDCM bit loop, where the exit step \(k_e\) is distinct from the
codebook size \(K\); \(k_e=8\) uses roughly 30\% of the bit-carrying steps and
\(k_e=14\) roughly 52\%.

\subsection{4K Perceptual Single-Example Result}
\label{app:pid-results}

\begin{table}[t]
\centering
\caption{PiD-DDCM 4K perceptual decode on one DIV8K crop (single-example diagnostic).  PiD
improves LPIPS and no-reference perceptual scores but lowers PSNR.  This is a
perception-distortion tradeoff, not a fidelity improvement.}
\label{tab:pid-smoke}
\small
\setlength{\tabcolsep}{2pt}
\begin{tabular}{@{}L{0.24\linewidth}C{0.12\linewidth}C{0.10\linewidth}C{0.11\linewidth}C{0.12\linewidth}C{0.10\linewidth}@{}}
\toprule
Method & BPP\(_{4K}\) & PSNR & LPIPS & MUSIQ & NIQE \\
\midrule
Bicubic \(4\times\) & .00270 & 20.79 & .485 & 25.22 & 8.31 \\
Lanczos \(4\times\) & .00270 & 20.70 & .464 & 25.41 & 8.79 \\
\rowcolor{aaaihighlight}
PiD full & .00270 & 18.18 & .379 & 53.70 & 4.39 \\
Bicubic \(k_e{=}8\) & .00103 & 18.13 & .849 & 23.88 & 8.15 \\
\rowcolor{aaaihighlight}
PiD \(k_e{=}8\) & .00103 & 17.52 & .423 & 61.63 & 3.78 \\
Bicubic \(k_e{=}14\) & .00180 & 20.10 & .614 & 23.92 & 7.68 \\
\rowcolor{aaaihighlight}
PiD \(k_e{=}14\) & .00180 & 18.09 & .376 & 58.75 & 3.83 \\
\bottomrule
\end{tabular}
\end{table}

This single-example diagnostic isolates the perception--distortion tradeoff. PiD variants score better
on perceptual metrics: MUSIQ rises from about 25 for classical SR to 54--62, NIQE drops
from about 8 to about 4, and LPIPS improves from the 0.46--0.85 range to the
0.38--0.42 range.  PSNR moves the other way, falling from about 20--21 dB for
the cascade baselines to about 17.5--18.2 dB for PiD.  The \(k_e{=}8\) row even
scores higher MUSIQ than the source crop itself, indicating that the decoder
can hallucinate sharper-than-source texture.

\subsection{Same-Resolution Control and Early Exit}
\label{app:pid-control}

A 1024-domain control sweep confirms that PiD is not a generic decoder swap.
Downsampling the PiD \(4\times\) output back to 1024 and comparing against the
1024 input leaves PSNR and MS-SSIM roughly flat, while LPIPS is worse than the
VAE decode.  The perceptual gain in Table~\ref{tab:pid-smoke} is therefore tied
to the true \(4\times\) SR setting.

The same sweep clarifies when PiD helps.  At aggressive early exit \(k_e=8\), PiD
repairs a heavily truncated latent and gives a PSNR lift of about 1.3--1.5 dB on
the 1024 control images, with large LPIPS gains on several images.  As \(k_e\)
increases and the VAE baseline becomes more faithful, this advantage erodes; by
\(k_e=20\), PiD often worsens LPIPS because it adds generative texture on top of an
already adequate reconstruction.

\subsection{Low-Rank Pixel Correction Probe}
\label{app:pid-lowrank}

We also tested a possible route from perceptual decoding toward a real codec
gain: add a structured low-rank endpoint correction in pixel space before the
PiD decode.  For each \(16\times16\times3\) patch \(p\), the residual between the
ground truth and the clean decoder prediction is
\[
r_p=x_{0,p}-\hat{x}_{0,p} ,
\]
and a rank-\(r\) basis \(A_p\in\mathbb{R}^{768\times r}\) defines the endpoint
correction subspace
\begin{equation}
r_p^\parallel = A_pA_p^\top r_p .
\label{eq:pid-lowrank-proj}
\end{equation}
This correction differs in kind from injecting low-rank SDE noise: it modifies
the clean endpoint channel, leaving the generative noise channel white and
avoiding an out-of-distribution sampler.

\begin{table}[t]
\centering
\caption{Low-rank residual energy captured on three Kodak images.  PCA is an
oracle upper bound fitted to the evaluation residual; DCT is a free structured
basis; random is the NCS-style baseline.  The probe is unquantized and is not
yet a real-bit codec result.}
\label{tab:pid-lowrank}
\small
\setlength{\tabcolsep}{3pt}
\begin{tabular}{@{}L{0.31\linewidth}C{0.12\linewidth}C{0.12\linewidth}C{0.12\linewidth}C{0.12\linewidth}@{}}
\toprule
Basis & \(r=8\) & \(r=32\) & \(r=64\) & \(r=128\) \\
\midrule
PiD / PCA oracle & .72 & .91 & .96 & .98 \\
PiD / DCT & .59 & .79 & .88 & .94 \\
PiD / random & .01 & .04 & .08 & .17 \\
VAE / PCA oracle & .76 & .94 & .99 & 1.00 \\
\bottomrule
\end{tabular}
\end{table}

Table~\ref{tab:pid-lowrank} shows that structured bases capture far more residual
energy than random atoms; oracle PCA at \(r=32\) captures about 91\% of the
PiD residual energy.  The caveat is that this remains an upper-bound diagnostic.
The PCA basis is not deployable without training or transmission, the projection
is real-valued and unquantized, and no true RD-P curve has been demonstrated.
The low-rank branch is therefore left as follow-up work rather than claimed as a result
of this paper.

\subsection{Hallucination Caveat}
\label{app:pid-caveat}

PiD-DDCM should be read as a perceptual reconstruction demo, not as a
fidelity-preserving codec result.  The generated textures can look plausible and
score better under LPIPS, MUSIQ, and NIQE while still departing from the
original image.  The risk is most acute at very low BPP, where the compressed
latent cannot uniquely determine fine detail.  For forensic, scientific,
medical, or evidentiary use, the conservative decoder or a fidelity-oriented
metric should remain the primary evaluation.  A video extension would also need
a temporal pixel-diffusion decoder, since independent per-frame PiD decoding
risks flicker and drift.

\end{document}